%% file: main.tex
\documentclass[10pt,twocolumn,letterpaper]{article}

\usepackage{cvpr}              
\usepackage{kotex}
\usepackage{tikz}
\usepackage{comment}
\usepackage{color}

\usepackage{enumitem}
\usepackage{graphicx}
\usepackage{amsmath}
\usepackage{booktabs}
\usepackage{mathtools}
\usepackage{url}
\usepackage{comment}
\usepackage{color}
\usepackage{graphicx,setspace}
\usepackage{multirow, array, booktabs,makecell,xspace,bm}
\usepackage{pifont}
\usepackage{amsfonts}
\usepackage{arydshln}
\usepackage{dsfont}

\usepackage{stackengine}

\usepackage{scalerel}
\usepackage{esdiff} 

\usepackage{bbm}

\usepackage{algorithm}
\usepackage{algpseudocode}
\algnewcommand{\LineComment}[1]{\State \# #1} 

\newcommand{\cmark}{\ding{51}}%

\def\maketitlesupplementary
   {
   \newpage
       \twocolumn[
        \centering
        \Large
        \textbf{\thetitle\\--\textit{Supplementary Material}--}\\
        \vspace{1.5em}
       ] 
   }


\definecolor{cvprblue}{rgb}{0.21,0.49,0.74}
\usepackage[pagebackref,breaklinks,colorlinks,citecolor=cvprblue]{hyperref}

\def\paperID{1966} 
\def\confName{CVPR}
\def\confYear{2024}

\title{Learning to Visually Localize Sound Sources\\from Mixtures without Prior Source Knowledge}

\author{
Dongjin Kim$^{1,}$\thanks{Equal contribution} \qquad
Sung Jin Um$^{1,}$\footnotemark[1] \qquad
Sangmin Lee$^{2,}$\thanks{Corresponding author} \qquad
Jung Uk Kim$^{1,}$\footnotemark[2]\\
$^{1}$Kyung Hee University \qquad
$^{2}$University of Illinois Urbana-Champaign\\
{\tt\small \{rlaehdwls310, sungzin1, ju.kim\}@khu.ac.kr, sangminl@illinois.edu}
}

\begin{document}

\maketitle
\input{sec/0_abstract}

\input{sec/1_intro}
\input{sec/2_related_work}
\input{sec/3_method}
\input{sec/4_experiments}

\section{Conclusion}
In this paper, we propose an innovative approach to multi-sound source localization that does not rely on prior knowledge. The core of our method is the Iterative Object Identification (IOI) module, which effectively identifies sound-making objects through iterative processes. Our object similarity-aware clustering (OSC) loss function successfully guides the IOI module not only to merge regions associated with the same object but also to discern distinct objects from the background. We believe that our approach, the iterative object identification method enhances accuracy and has diverse practical applications.\\

\noindent{\textbf{Acknowledgements.}}
This work was conducted by CARAI grant funded by DAPA and ADD (UD230017TD), and supported in part by the NRF grant funded by the Korea government (MSIT) (No. RS-2023-00252391) and by IITP grant (No. 2022-0-00124), IITP grant (IITP-2023-RS-2023-00266615), IITP grant (No. RS-2022-00155911) and by the MSIT (Ministry of Science and ICT), Korea, under the National Program for Excellence in SW (2023-0-00042) supervised by the IITP in 2024.

{
    \small
    \bibliographystyle{ieeenat_fullname}
    \bibliography{main}
}


\clearpage

\setcounter{section}{0}
\setcounter{figure}{0}
\setcounter{table}{0}
\setcounter{equation}{0}
\pagenumbering{gobble}


\maketitlesupplementary


\noindent This manuscript provides additional implementation details and additional results of the proposed method. In Section 1, we elaborate on the additional implementation details of our method. Section 2 presents additional experimental results to show the effectiveness of the Iterative Object Identification (IOI) module and object similarity-aware clustering (OSC) loss. Moreover, Section 3 shows additional visualization results. Note that [PXX] indicates the reference in the main paper.

\section{Additional Implementation Details}

We utilize the ResNet-18 [P13] for the audio encoder, as mentioned in the main paper. Since the audio spectrogram has only one channel, we modify the first convolution layer of the encoder to have an input channel of 1 and an output channel of 64, utilizing a kernel size of 7, stride of 2, and padding of 3. Additionally, we employ the Adam optimizer, setting the parameters $(\beta_1, \beta_2)$ to (0.9, 0.999), which are the standard values for Adam. For the hyperparameter $\theta$ and $\omega$, mentioned in Section 3.2, adopt the values 0.65 and 0.03, respectively, following [P8].

\section{Additional Experiments}

\begin{figure}[t]
    \begin{minipage}[b]{1.0\linewidth}
			\centering
			\vspace{-0.3cm}
				\centerline{\includegraphics[width=8.7cm]{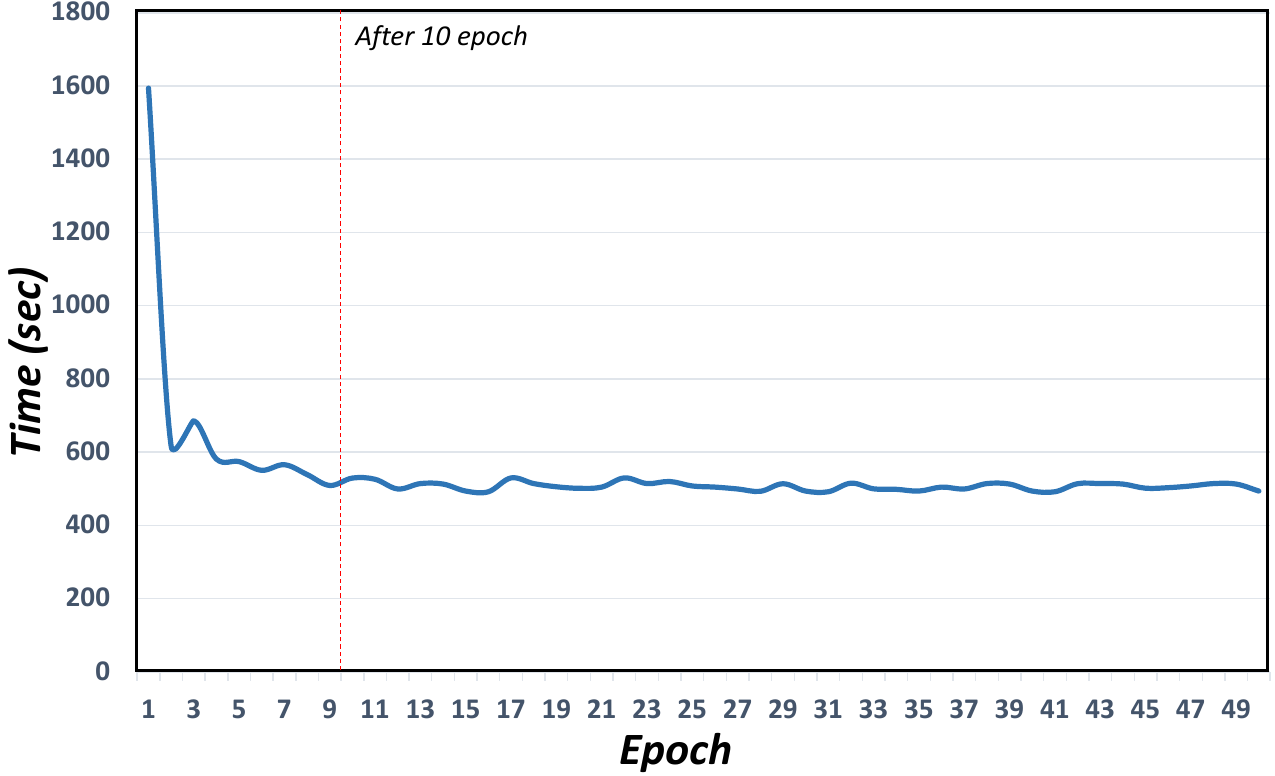}}
			\end{minipage}
			\vspace{-0.7cm}
			\caption{Visualization of training time per epoch of our model. After epoch 10 (red line), the training time converges to about 500 seconds/epoch.}
			\vspace{-0.5cm}
   \label{fig:time}
		\end{figure}

\noindent {\textbf{Training Time per Epoch in Training Phase.}}
Since our approach adopts an iterative method, we explored how training duration varies across epochs, as illustrated in Figure \ref{fig:time}. Initially, in the first epoch, the processing time is high at 1,594 seconds per epoch. However, with more epochs, this time significantly drops and stabilizes around 500 seconds per epoch. This trend suggests that our method becomes more computationally efficient over time by reducing unnecessary iterations, focusing on the effective steps for localizing sound-making objects. \\

\noindent {\textbf{Comparison of our method with baseline (single localization applied after separation).}}
We present an additional experiment to validate the robustness of our approach for localizing sound sources from mixtures by comparing it with a baseline method which is first to perform audio source separation on the mixture and then apply single sound source localization to each segregated audio element. On MUSIC-Duet, our method is superior to the baseline (widely used audio separation model [\textcolor{cyan}{1}] followed by single SSL), showing CAP (22.4$\rightarrow$52.1), PIAP (44.8$\rightarrow$72.5), CIoU@0.3 (29.8$\rightarrow$38.6), and AUC (23.6$\rightarrow$30.1). \\

\noindent {\textbf{Impact of audio component.}}
We conducted the visual-only experiment on the MUSIC-Duet data set to investigate the significance of the audio component. Adding an audio component enhances performance across all metrics: CAP (20.5$\rightarrow$52.1), PIAP (31.4$\rightarrow$72.5), CIoU@0.3 (26.1$\rightarrow$38.6), and AUC (21.2$\rightarrow$30.1). \\

\section{Additional Visualization Results}
\begin{figure*}[t]
    \begin{minipage}[b]{1.0\linewidth}
				\centering
				\centerline{\includegraphics[width=14cm]{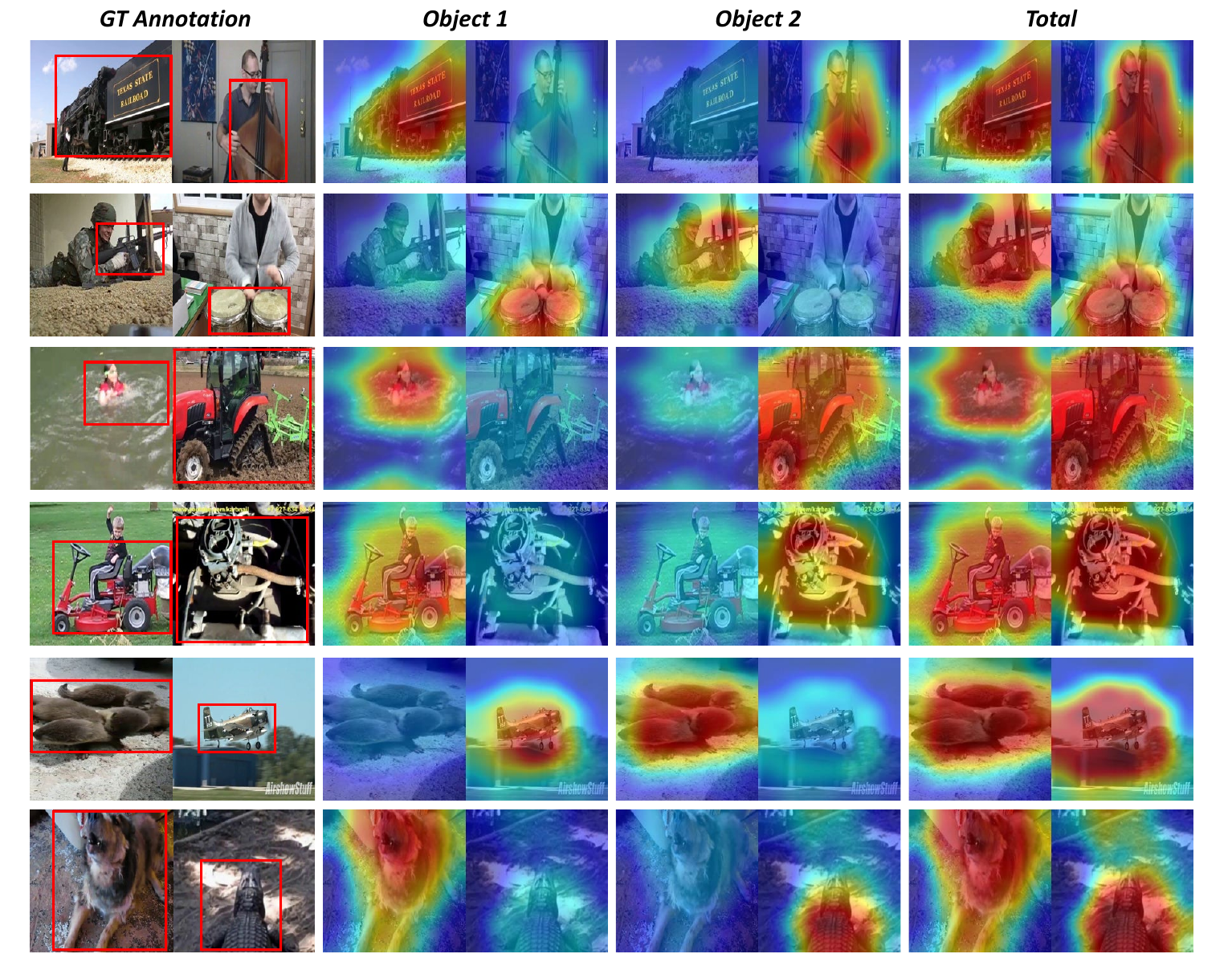}}
			\end{minipage}
			\vspace{-0.69cm}
			\caption{Additional visualization results for VGGSound-Duet test set (two objects). `Object $k$' is identified by our model without any prior knowledge.}
   \label{fig:duet}
\end{figure*}

\begin{figure*}[t]
    \begin{minipage}[b]{1.0\linewidth}
				\centering
				\centerline{\includegraphics[width=18cm]{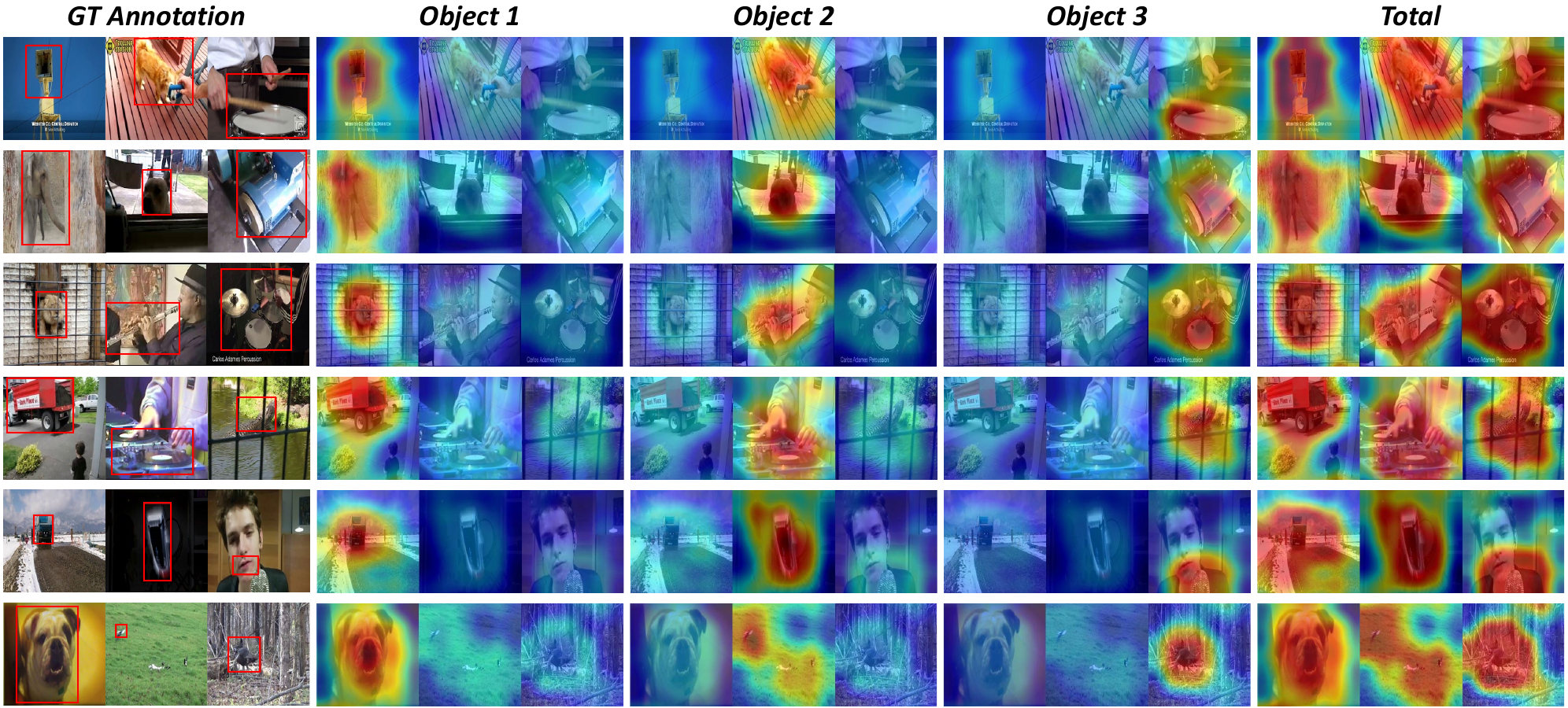}}
			\end{minipage}
			\vspace{-0.69cm}
			\caption{Additional visualization results for VGGSound-Trio test set (three objects). `Object $k$' is identified by our model without any prior knowledge.}
   \vspace{0.2cm}
   \label{fig:trio}
\end{figure*}

\begin{figure*}[t]
    \begin{minipage}[b]{1.0\linewidth}
				\centering
				\centerline{\includegraphics[width=12cm]{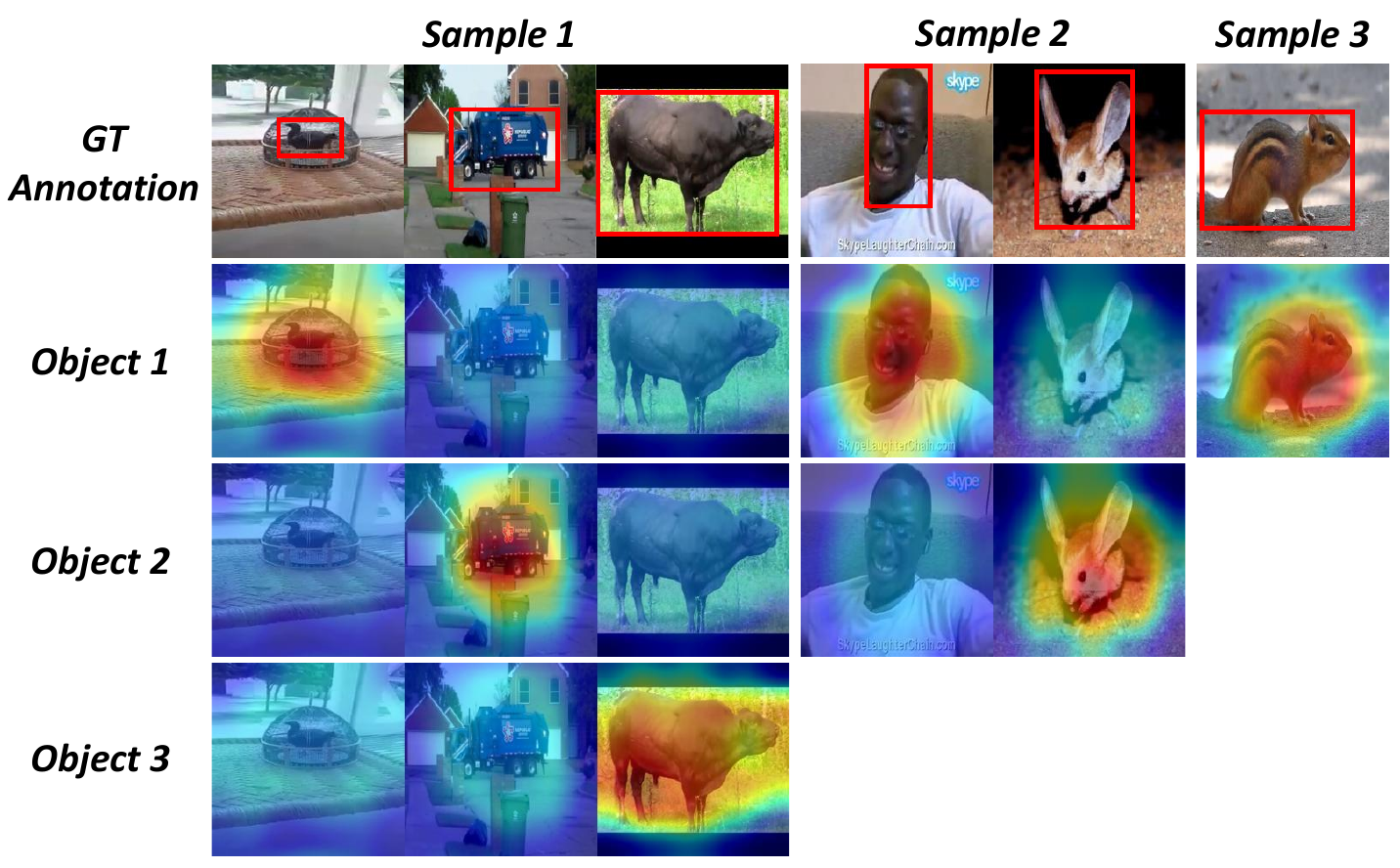}}
			\end{minipage}
			\vspace{-0.69cm}
			\caption{Additional visualization results for VGGSound-Mixed test set (mixed objects). `Object $k$' is identified by our model without any prior knowledge.}
			\vspace{-0.12cm}
   \label{fig:mixed}
\end{figure*}

\noindent {\textbf{Visualization Results on VGGSound-Duet, Trio and Mixed Dataset.}}  We present additional visualization results of our method to demonstrate its efficacy in differentiating objects in scenarios with various source mixtures utilizing a VGGSound-Duet, Trio, and VGGSound-Mixed test set. VGGSound-Trio test set is comprised mixture of three sound sources from VGGSound-Single [P7], as guided by [P26], and VGGSound-Mixed test set.  Our IOI module is adept at repeatedly detecting and distinguishing sound-making objects within audio-visual scenes, resulting in highly accurate and detailed localization maps. These visualizations, as depicted in Figures \ref{fig:duet}, \ref{fig:trio} and \ref{fig:mixed}, demonstrate the accuracy of our model in individual object localization and total map estimation, reflecting a deep understanding of the complex audio-visual landscape.\\

\newpage

\begin{figure*}[t]
    \begin{minipage}[b]{1.0\linewidth}
				\centering
				\centerline{\includegraphics[width=12cm]{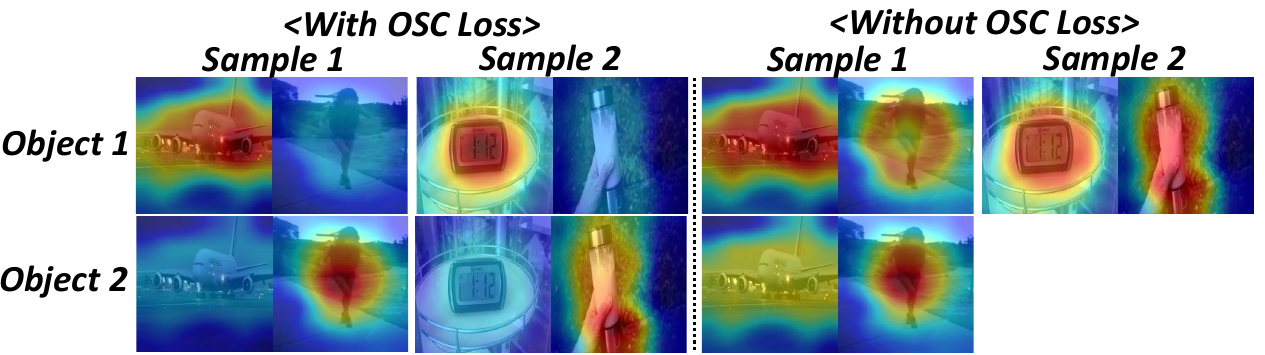}}
			\end{minipage}
			\vspace{-0.69cm}
			\caption{Additional visualization results with/without OSC loss. `Object $k$' is identified by our model without any prior knowledge.}
   \label{fig:osc}
\end{figure*}

\noindent {\textbf{Visualization Results without OSC loss.}} Based on the ablation study on the effect of the proposed losses within our manuscript, we present visualization results for scenarios with and without the OSC (Object Similarity-aware Clustering) loss. This comparison is based on two samples used in Figure 5 of the manuscript. In Figure \ref{fig:osc}, the results demonstrate that the use of OSC loss leads to better performance in separating objects in mixtures. Consequently, it can be observed that the ability of our model to distinguish between objects is enhanced through the incorporation of the OSC loss.\\

\noindent {\textbf{Video Demo.}} We provide video materials that offer a more in-depth explanation of our method for localizing sound-making objects in complex environments. These videos demonstrate the real-time applicability and robustness of our approach under various conditions. We provide results of our method with some examples from the VGGSound-Duet dataset. Please see video in our  \href{https://github.com/VisualAIKHU/NoPrior_MultiSSL}{\color{black}{official repository}}.


\clearpage

\section*{References}
\small [1] Subakan et al. Attention is all you need in speech separation. In \textit{ICASSP}, 2021. \\
\small [P7] Honglie Chen, Weidi Xie, Andrea Vedaldi, and Andrew Zisserman. Vggsound: A large-scale audio-visual dataset. In \textit{ICASSP}, 2020. \\
\small [P8] Honglie Chen, Weidi Xie, Triantafyllos Afouras, Arsha Nagrani, Andrea Vedaldi, and Andrew Zisserman. Localizing visual sounds the hard way. In \textit{CVPR}, 2021.\\
\small [P13] Kaiming He, Xiangyu Zhang, Shaoqing Ren, and Jian Sun. Deep residual learning for image recognition. In \textit{CVPR}, 2016. \\
\small [P26] Shentong Mo and Yapeng Tian. Audio-visual grouping network for sound localization from mixtures. In \textit{CVPR}, 2023. \\

\end{document}


\maketitle

\noindent This manuscript provides additional implementation details and additional results of the proposed method. In Section 1, we elaborate on the additional implementation details of our method. Section 2 presents additional experimental results to show the effectiveness of the Iterative Object Identification (IOI) module and object similarity-aware clustering (OSC) loss. Moreover, Section 3 shows additional visualization results. Note that [PXX] indicates the reference in the main paper.

\section{Additional Implementation Details}

We utilize the ResNet-18 [P13] for the audio encoder, as mentioned in the main paper. Since the audio spectrogram has only one channel, we modify the first convolution layer of the encoder to have an input channel of 1 and an output channel of 64, utilizing a kernel size of 7, stride of 2, and padding of 3. Additionally, we employ the Adam optimizer, setting the parameters $(\beta_1, \beta_2)$ to (0.9, 0.999), which are the standard values for Adam. For the hyperparameter $\theta$ and $\omega$, mentioned in Section 3.2, adopt the values 0.65 and 0.03, respectively, following [P8].


\section{Additional Experiments}

\begin{figure}[t]
    \begin{minipage}[b]{1.0\linewidth}
			\centering
			\vspace{-0.3cm}
				\centerline{\includegraphics[width=8.7cm]{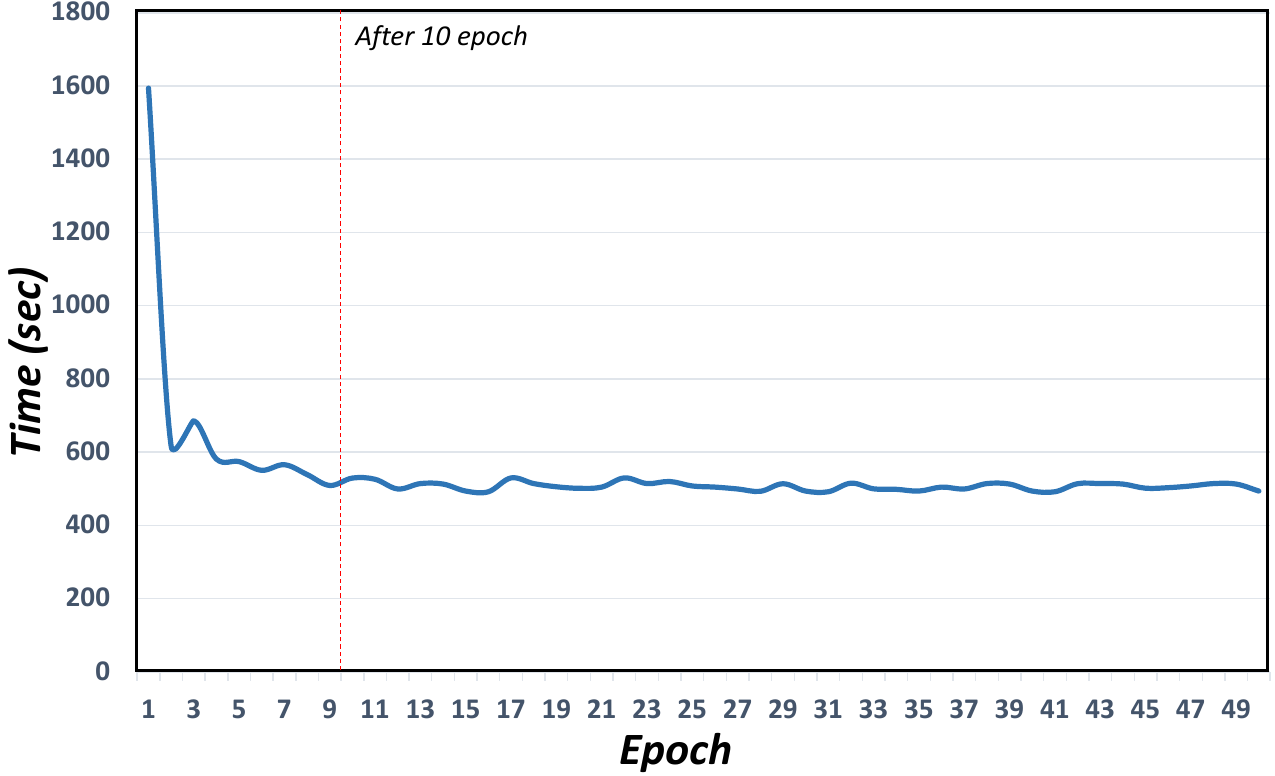}}
			\end{minipage}
			\vspace{-0.7cm}
			\caption{Visualization of training time per epoch of our model. After epoch 10 (red line), the training time converges to about 500 seconds/epoch.}
			\vspace{-0.5cm}
   \label{fig:time}
		\end{figure}

\noindent {\textbf{Training Time per Epoch in Training Phase.}}
Since our approach adopts an iterative method, we investigated how training time changes over epochs. The results are shown in Figure \ref{fig:time}. In the early epochs of the training phase (\textit{i.e.,} 1 epoch), the processing time is relatively high, showing, 1,594 seconds/epoch. However, as the number of epochs increases, the training time significantly decreases and converges to approximately 500 seconds/epoch. This indicates that, as the number of epochs increases, unnecessary iteration steps decrease. In other words, our proposed method becomes computationally more efficient over time by utilizing only effective iteration steps for the localization of sound-making objects. \\

\noindent {\textbf{Comparison of our method with baseline (single localization applied after separation).}}
We present an additional experiment to validate the robustness of our approach for localizing sound sources from mixtures by comparing it with a baseline method which is first to perform audio source separation on the mixture and then apply single sound source localization to each segregated audio element. On MUSIC-Duet, our method is superior to the baseline (widely used audio separation model [\textcolor{cyan}{1}] followed by single SSL), showing CAP (22.4$\rightarrow$52.1), PIAP (44.8$\rightarrow$72.5), CIoU@0.3 (29.8$\rightarrow$38.6), and AUC (23.6$\rightarrow$30.1). \\

\noindent {\textbf{Impact of audio component.}}
We conducted the visual-only experiment on the MUSIC-Duet data set to investigate the significance of the audio component. Adding an audio component enhances performance across all metrics: CAP (20.5$\rightarrow$52.1), PIAP (31.4$\rightarrow$72.5), CIoU@0.3 (26.1$\rightarrow$38.6), and AUC (21.2$\rightarrow$30.1). \\


\begin{figure*}[t]
    \begin{minipage}[b]{1.0\linewidth}
				\centering
				\centerline{\includegraphics[width=14cm]{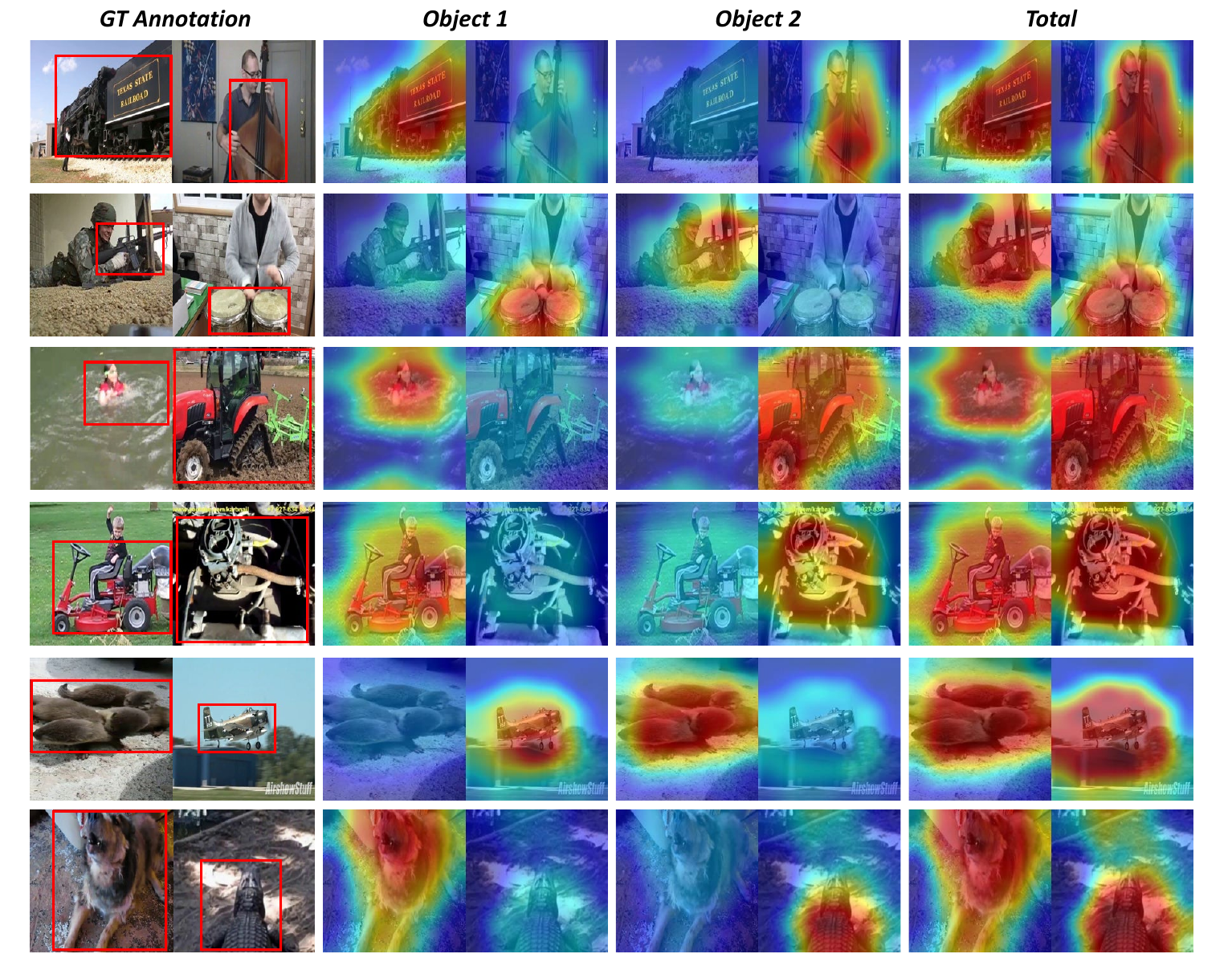}}
			\end{minipage}
			\vspace{-0.69cm}
			\caption{Additional visualization results for VGGSound-Duet test set (two objects). `Object $k$' is identified by our model without any prior knowledge.}
			\vspace{-0.12cm}
   \label{fig:duet}
\end{figure*}

\begin{figure*}[t]
    \begin{minipage}[b]{1.0\linewidth}
				\centering
				\centerline{\includegraphics[width=18cm]{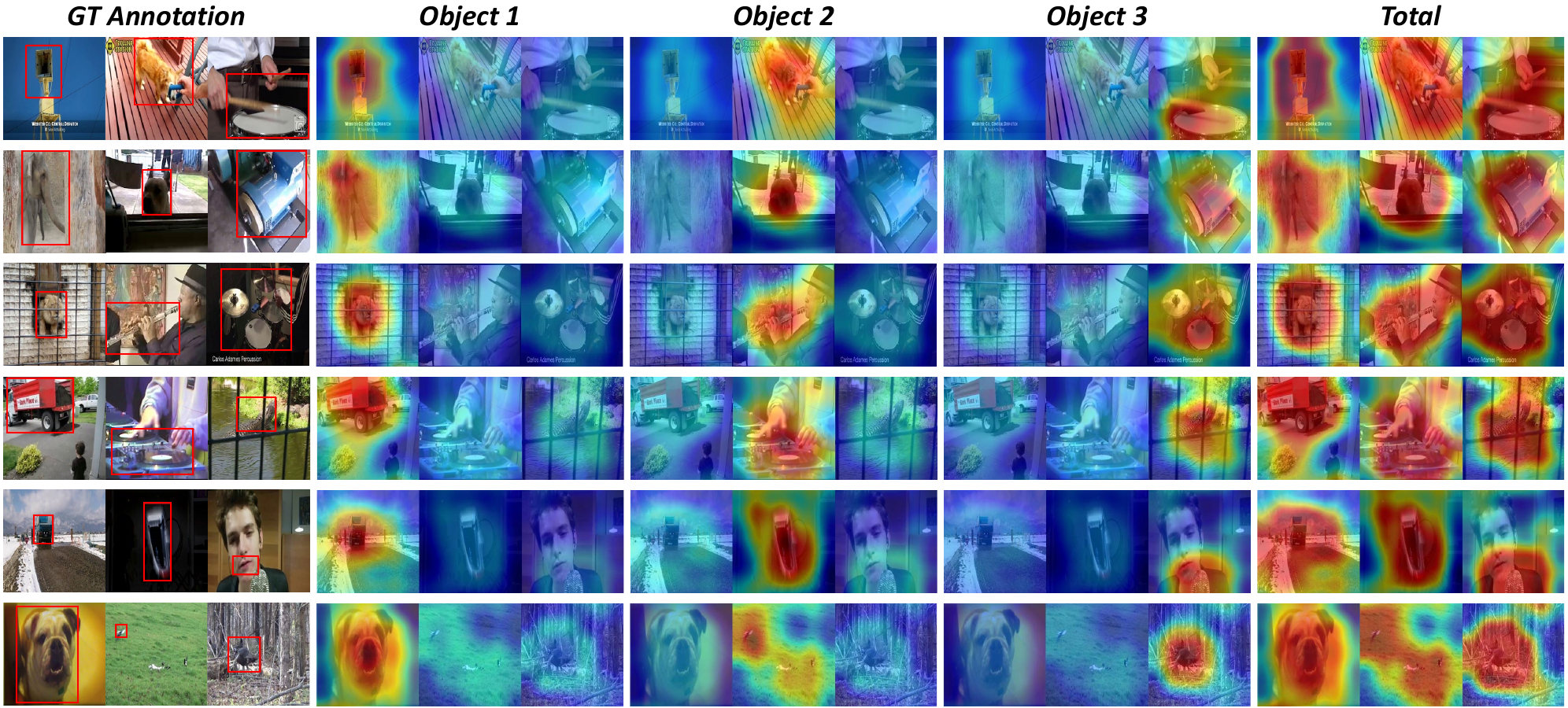}}
			\end{minipage}
			\vspace{-0.69cm}
			\caption{Additional visualization results for VGGSound-Trio test set (three objects). `Object $k$' is identified by our model without any prior knowledge.}
			\vspace{-0.12cm}
   \label{fig:trio}
\end{figure*}

\begin{figure*}[t]
    \begin{minipage}[b]{1.0\linewidth}
				\centering
				\centerline{\includegraphics[width=12cm]{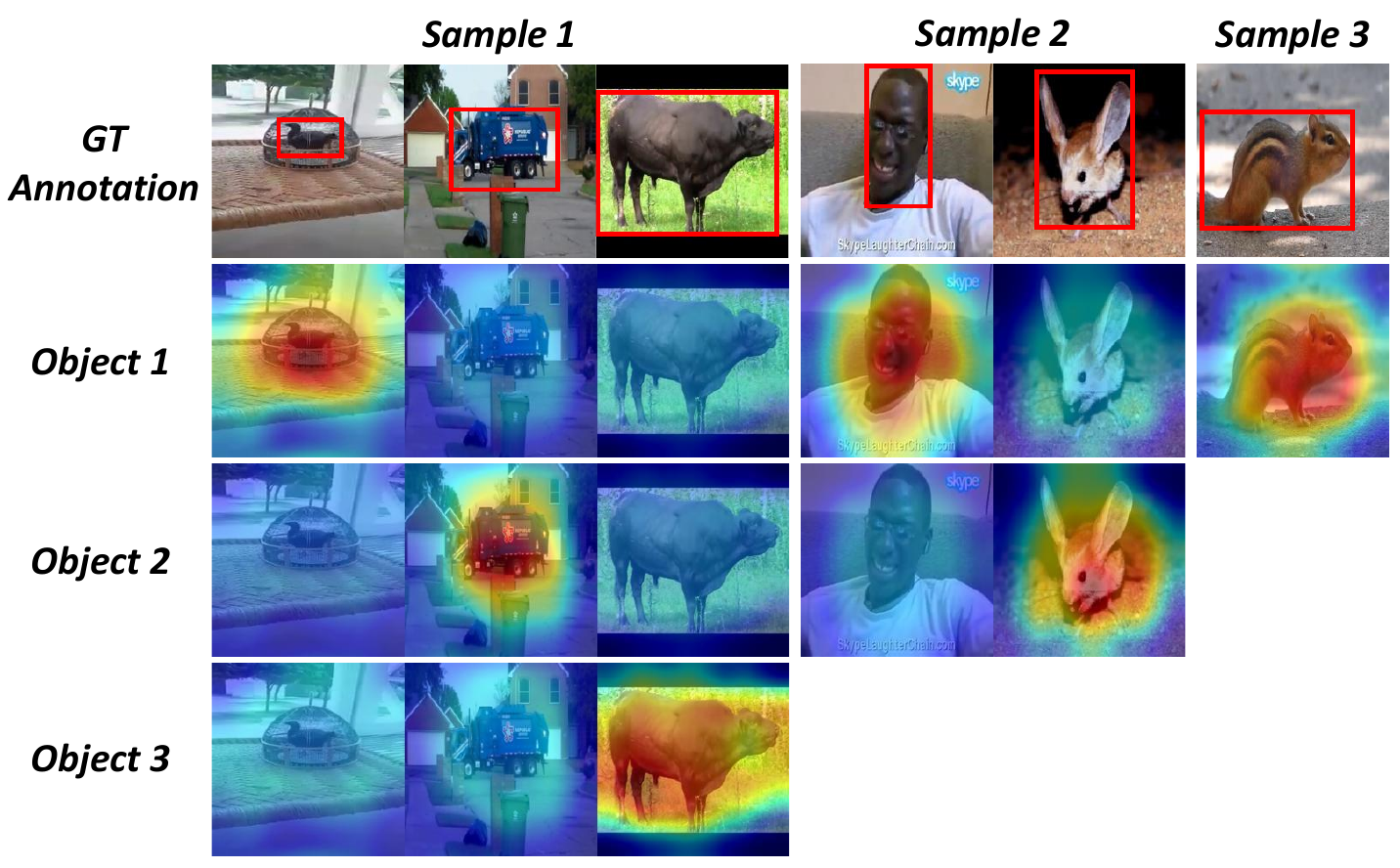}}
			\end{minipage}
			\vspace{-0.69cm}
			\caption{Additional visualization results for VGGSound-Mixed test set (mixed objects). `Object $k$' is identified by our model without any prior knowledge.}
			\vspace{-0.12cm}
   \label{fig:mixed}
\end{figure*}

\begin{figure*}[t]
    \begin{minipage}[b]{1.0\linewidth}
				\centering
				\centerline{\includegraphics[width=12cm]{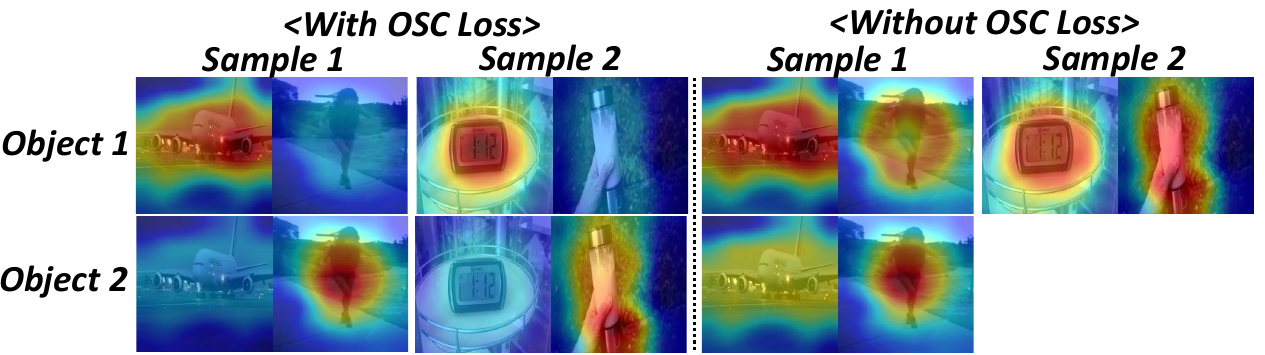}}
			\end{minipage}
			\vspace{-0.69cm}
			\caption{Additional visualization results with/without OSC loss. `Object $k$' is identified by our model without any prior knowledge.}
			\vspace{-0.12cm}
   \label{fig:osc}
\end{figure*}


\section{Additional Visualization Results}
\vspace{-0.1cm}
\noindent {\textbf{Visualization Results on VGGSound-Duet, Trio and Mixed Dataset.}}  We present additional visualization results of our method to demonstrate its efficacy in differentiating objects in scenarios with various source mixtures utilizing a VGGSound-Duet, Trio, and VGGSound-Mixed test set. VGGSound-Trio test set is comprised mixture of three sound sources from VGGSound-Single [P7], as guided by [P26], and VGGSound-Mixed test set.  Our IOI module is adept at repeatedly detecting and distinguishing sound-making objects within audio-visual scenes, resulting in highly accurate and detailed localization maps. These visualizations, as depicted in Figures \ref{fig:duet}, \ref{fig:trio} and \ref{fig:mixed}, demonstrate the accuracy of our model in individual object localization and total map estimation, reflecting a deep understanding of the complex audio-visual landscape.\\

\noindent {\textbf{Visualization Results without OSC loss.}} Based on the ablation study on the effect of the proposed losses within our manuscript, we present visualization results for scenarios with and without the OSC (Object Similarity-aware Clustering) loss. This comparison is based on two samples used in Figure 5 of the manuscript. In Figure \ref{fig:osc}, the results demonstrate that the use of OSC loss leads to better performance in separating objects in mixtures. Consequently, it can be observed that the ability of our model to distinguish between objects is enhanced through the incorporation of the OSC loss.\\

\noindent {\textbf{Video Demo.}} We provide video materials that offer a more in-depth explanation of our method for localizing sound-making objects in complex environments. These videos demonstrate the real-time applicability and robustness of our approach under various conditions. We provide results of our method with some examples from the VGGSound-Duet dataset. Please see ``CVPR2024\_Submission\_PaperID1966\_Supp\_Video.mp4.''

\clearpage

\section*{References}
\small [1] Subakan et al. Attention is all you need in speech separation. In \textit{ICASSP}, 2021. \\
\small [P7] Honglie Chen, Weidi Xie, Andrea Vedaldi, and Andrew Zisserman. Vggsound: A large-scale audio-visual dataset. In \textit{ICASSP}, 2020. \\
\small [P8] Honglie Chen, Weidi Xie, Triantafyllos Afouras, Arsha Nagrani, Andrea Vedaldi, and Andrew Zisserman. Localizing visual sounds the hard way. In \textit{CVPR}, 2021.\\
\small [P13] Kaiming He, Xiangyu Zhang, Shaoqing Ren, and Jian Sun. Deep residual learning for image recognition. In \textit{CVPR}, 2016. \\
\small [P26] Shentong Mo and Yapeng Tian. Audio-visual grouping network for sound localization from mixtures. In \textit{CVPR}, 2023. \\

%% file: sec/0_abstract.tex
\begin{abstract}
The goal of the multi-sound source localization task is to localize sound sources from the mixture individually. While recent multi-sound source localization methods have shown improved performance, they face challenges due to their reliance on prior information about the number of objects to be separated. In this paper, to overcome this limitation, we present a novel multi-sound source localization method that can perform localization without prior knowledge of the number of sound sources. To achieve this goal, we propose an iterative object identification (IOI) module, which can recognize sound-making objects in an iterative manner. After finding the regions of sound-making objects, we devise object similarity-aware clustering (OSC) loss to guide the IOI module to effectively combine regions of the same object but also distinguish between different objects and backgrounds. It enables our method to perform accurate localization of sound-making objects without any prior knowledge. Extensive experimental results on the MUSIC and VGGSound benchmarks show the significant performance improvements of the proposed method over the existing methods for both single and multi-source. Our code is available at: \href{https://github.com/VisualAIKHU/NoPrior_MultiSSL}{\color{magenta}{https://github.com/VisualAIKHU/NoPrior\_MultiSSL}}.
\end{abstract}

%% file: sec/1_intro.tex
\vspace{-0.45cm}
\section{Introduction}
\label{sec:intro}

Humans naturally perceive various sounds and identify the origins of those sounds by using both visual sense (eyes) and auditory sense (ears) \cite{hawley1999speech}. The sound source localization task aims to mimic the ability of humans to correlate auditory cues with visual information in order to identify sound-making objects. Due to this property, sound source localization is closely related to various real-world applications, including unmanned aerial vehicles \cite{hoshiba2017_uav, salvati2019_uav}, robotics \cite{li2016robotics,rascon2017robotics}, and speaker source localization \cite{sayoud2011speaker, busso2005smart_speaker}.

The sound source localization task can be divided into two categories: (1) single sound source localization and (2) multi-sound source localization. The single sound source localization task \cite{s_CVPR2018_Senocak, s_CVPR2021_lvs, s_hard_positive_mining, s_iterative2023, s_um2023sira, s_sun2023learning, s_WACV2022_Shi, s_WACV2023_htf,s_momentum,s_flowgrad, s_WACV2023_Zhou, s_CVPR2022_ppsl, s_liu2022exploiting, s_senocak2023sound, s_song2022sspl} aims to find one source in a scene by utilizing cross-modal correlations \cite{s_CVPR2018_Senocak,s_crossmodal} between audio and visual cues. Various methods have been developed for the effective single sound source localization by introducing hard positive mining \cite{s_CVPR2021_lvs, s_hard_positive_mining}, iterative learning \cite{s_iterative2023}, feature regularization \cite{s_liu2022exploiting}, negative free learning \cite{s_song2022sspl}, false negative aware learning \cite{s_sun2023learning}, momentum target encoders \cite{s_momentum}, optical flow \cite{s_WACV2023_htf, s_flowgrad}, and spatial integration \cite{s_um2023sira}. However, these methods focus primarily on locating a single sound source, which can be challenging in real-world environments where multiple sounds are often mixed together.

\begin{figure}[t]
    \begin{minipage}[b]{1.0\linewidth}
    \vspace{-0.2cm}
	\centering
	\centerline{\includegraphics[width=8.4cm]{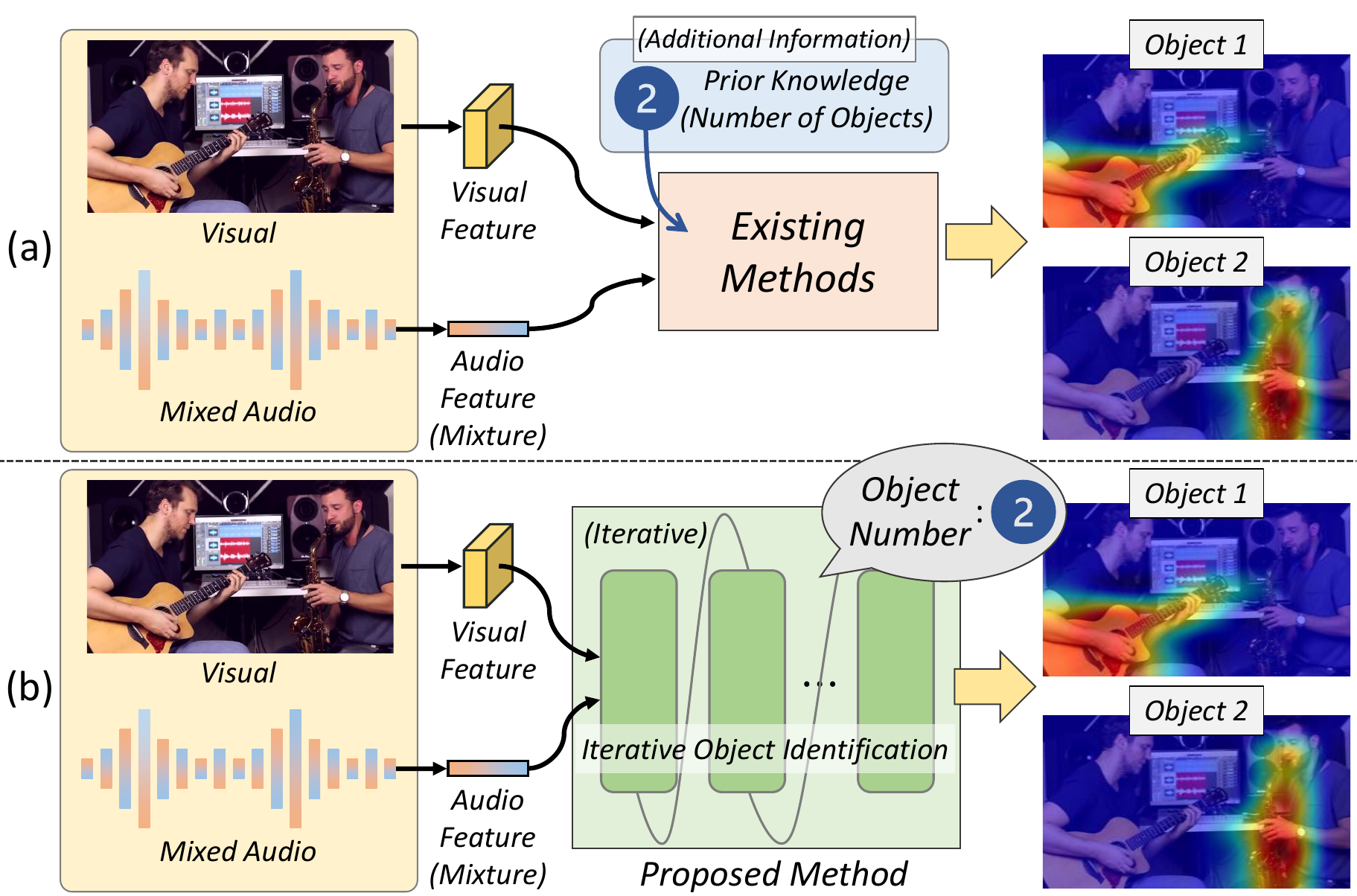}}
	\end{minipage}
        \vspace{-0.7cm}
	\caption{Conceptual comparison between (a) existing methods and (b) the proposed method. The existing methods require prior source knowledge of the number of sound-making objects. In contrast, our method can effectively localize multiple sound-making objects without the need for prior source knowledge.}
    \vspace{-0.6cm}
    \label{fig:1}
\end{figure}
In response to this challenge, several multi-sound source localization methods \cite{m_ECCV2020_Qian_coarsetofine,m_hu2020discriminative,m_hu2022mix,m_mo2023audio} have been developed. The main goal of multi-sound source localization is to separate and localize individual sources from complex mixtures containing different sounds, such as self-supervised audio-visual matching \cite{m_hu2020discriminative}, coarse-to-fine manner \cite{m_ECCV2020_Qian_coarsetofine}, contrastive random walker \cite{m_hu2022mix}, and audio-visual grouping network \cite{m_mo2023audio}. However, a limitation of existing methods is their reliance on prior information about the number of objects that need to be separated. As shown in Figure \ref{fig:1}(a), existing methods heavily rely on prior knowledge about the number of sound sources. If prior knowledge is incorrect, they are frequently failed to localize sound-making objects. Thus, they can only operate in constrained environments where prior source knowledge is available for sound source localization. Consequently, accurate localization of multi-sound sources becomes challenging when this prior knowledge is not available. In addition, since prior knowledge is generally not provided in real-world environments, it limits their applicability in practical scenarios.

To address the aforementioned challenges, we introduce a novel method for multi-sound source localization that can identify multiple sound sources without the need for prior knowledge of the number of sources. As shown in Figure \ref{fig:1}(b), our method can adapt to various numbers of sound sources by automatically recognizing the number of sound-making objects without relying on any prior knowledge. To this end, we propose an Iterative Object Identification (IOI) module. The goal of our IOI is to effectively automate object separation by repeatedly recognizing objects that make sounds from the mixtures in an iterative manner. By doing so, it continuously searches for sound-making regions until it determines that all relevant objects have been identified.

To achieve this goal, we provide guidance to the IOI module for effective sound-making object identification. First, in the iterative process, we identify regions that are considered to be foreground objects and repeat this process until there are no more regions considered as foreground objects. At each iteration step, regions previously considered as foreground objects are eliminated in the subsequent iteration. Next, we employ a clustering process to merge regions belonging to the same object based on the foreground regions. To effectively carry out this process, we devise an object similarity-aware clustering (OSC) loss. This loss not only guides the IOI module to combine regions of the same object but also aids in distinguishing between different objects and backgrounds. By doing so, the proposed framework is able to distinguish between various objects with distinct sounds through the iterative process.

Consequently, our method can precisely locate the sound-making objects without the need for prior source knowledge, and can effectively distinguish between multiple sound sources. As a result, our method shows significant improvements in sound source localization performance compared to existing methods. 

The major contributions of our paper are as follows:

\begin{itemize}
    \item We introduce Iterative Object Identification (IOI) module to adaptively localize multiple sound sources without any prior source knowledge. To the best of our knowledge, this is the first attempt to perform multi-sound source localization without knowing the number of sound sources.

    \item We propose Object Similarity-aware Clustering (OSC) loss to guide the IOI module to merge regions belonging to the same object during the iteration process, while distinguishing between different objects and the background.

    \item Experimental results on MUSIC and VGGSound datasets demonstrate the effectiveness of the proposed method for both single-/multi-source sound localization.
\end{itemize}
 

%% file: sec/2_related_work.tex
\section{Related Work}
\label{sec:related_works}

\subsection{Sound Source Localization}
The sound source localization focuses on finding the spatial location of sound sources within a video frame. It can be divided into two main streams: (1) single sound source localization and (2) multi-sound source localization.

\begin{figure*}[t]
    \begin{minipage}[b]{1.0\linewidth}
	\centering
    \vspace{-0.9cm}
	\centerline{\includegraphics[width=17cm]{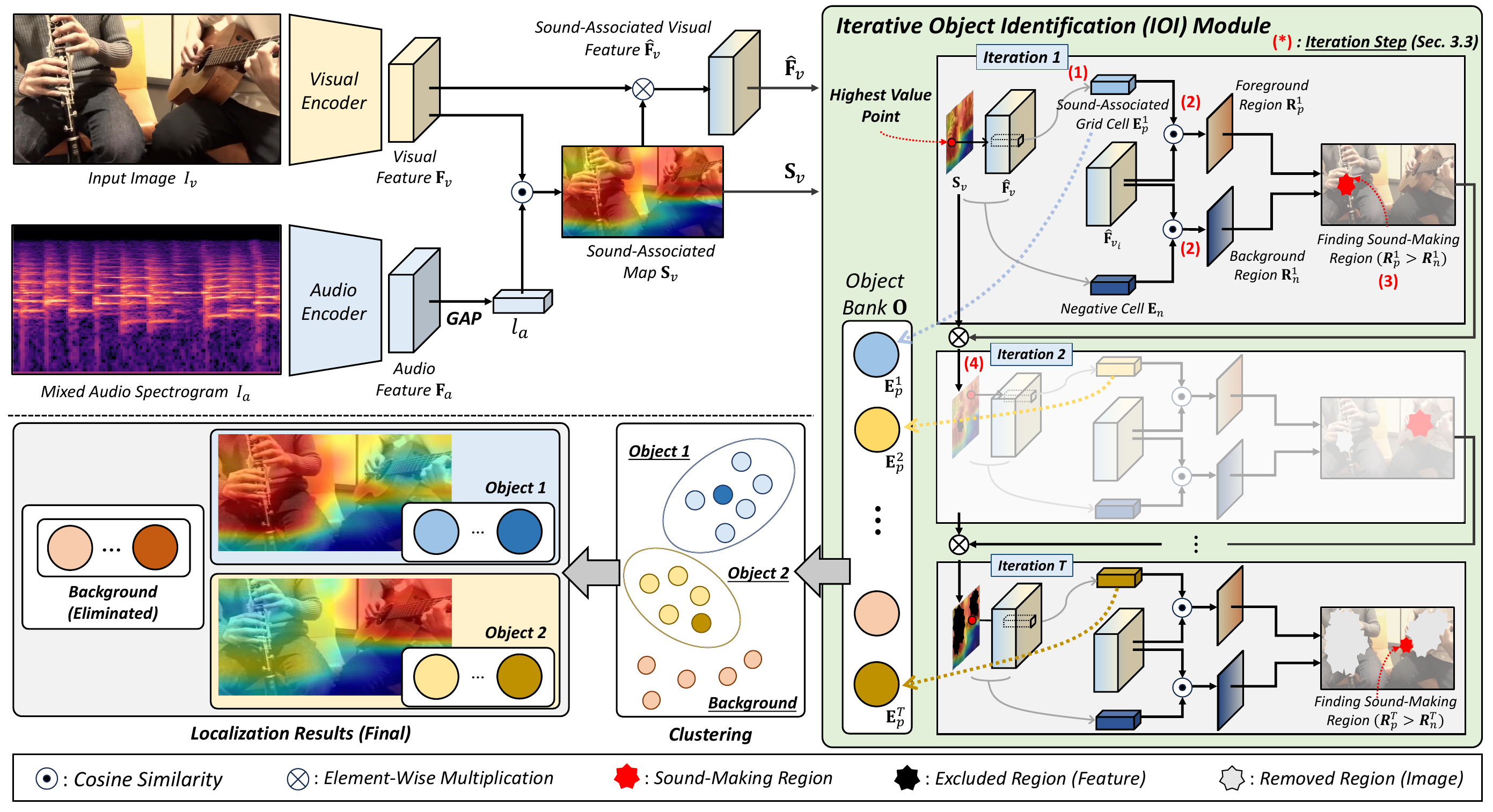}}
	\end{minipage}
    \vspace{-0.75cm}
    \caption{Network configuration of the proposed sound source localization framework. GAP denotes global average pooling.}
    \label{fig:2}
    \vspace{-0.4cm}
\end{figure*}

The single sound source localization \cite{s_CVPR2018_Senocak, s_CVPR2021_lvs, s_hard_positive_mining, s_iterative2023, s_um2023sira, s_sun2023learning, s_WACV2022_Shi, s_WACV2023_htf,s_chen2021exploring,s_momentum,s_flowgrad, s_WACV2023_Zhou, s_CVPR2022_ppsl, s_liu2022exploiting, s_senocak2023sound, s_song2022sspl} leverages the cross-modal interrelations between auditory and visual modalities to localize a sound-making object. Senocak \textit{et al.} \cite{s_CVPR2018_Senocak} introduced an unsupervised method to incorporate an audio-visual two-stream network with attention mechanisms. Chen \textit{et al.} \cite{s_CVPR2021_lvs} proposed a framework that automatically identifies hard samples through contrastive learning. Lin \textit{et al.} \cite{s_iterative2023} adopted iterative learning with pseudo-labels to refine the localization process. In addition, Fedorishin \textit{et al.} \cite{s_WACV2023_htf} and Singh \textit{et al.} \cite{s_flowgrad} leveraged optical flow as a prior and introduced a cross-attention mechanism over the relevant video frame. Sun \textit{et al.} \cite{s_sun2023learning} introduced the False Negative Aware Contrastive (FNAC) strategy to guide similar-looking samples to be more similar. Senocak \textit{et al.} \cite{s_senocak2023sound} improved localization accuracy through enhanced cross-modal semantic understanding. Um \textit{et al.} \cite{s_um2023sira} proposed the spatial knowledge integration of the audio-visual modalities, promoting recursive enhancement of localization. However, in real-world scenarios, they encounter limitations due to the frequent occurrence of mixed sounds from various sources.

Recent studies focus on the multi-sound source localization \cite{m_ECCV2020_Qian_coarsetofine,m_hu2020discriminative,m_hu2022mix,m_mo2023audio} to effectively find the location of sound sources from mixtures. Hu \textit{et al.} \cite{m_hu2020discriminative} introduced a technique that constructs a supervisory signal to differentiate the sound-making entities within a scene. Qian \textit{et al.} \cite{m_ECCV2020_Qian_coarsetofine} developed a two-stage audiovisual learning approach that separates audio and visual representations of various categories from scenes in a coarse-to-fine manner. Hu \textit{et al.} \cite{m_hu2022mix} made a graph model for robust multi-sound source localization. Mo \textit{et al.} \cite{m_mo2023audio} proposed a way to simultaneously extract and assimilate category-specific semantic features for each sound source from auditory and visual inputs.

However, the existing multi-sound source localization methods also exhibit a limitation in real-world scenarios. They heavily rely on prior knowledge about the number of sound sources to differentiate  between multi-sound sources. Thus, we propose a sound source localization framework that can autonomously recognize the number of objects within a scene and localize the objects without relying on annotations or prior knowledge.

\subsection{Iterative Methods in Computer Vision}
Iterative techniques have been applied to various research fields \cite{i_image_deblurring2,i_image_deblurring1,i_super_resolution,i_object_detection1,i_object_detection2,i_segmentation1,i_segmentation2,i_segmentation3,i_face_super,i_single_view_depth,i_sfm} because they provide a refining solutions through repetitive cycles. For example, the iterative methods have been applied to image deblurring and denoising tasks \cite{i_image_deblurring2,i_image_deblurring1}, showing significant improvements over traditional non-iterative techniques. In addition, for the mainstrean computer vision tasks, e.g., object detection and segmentation tasks, various works \cite{i_object_detection1,i_object_detection2,i_segmentation1,i_segmentation2,i_segmentation3} also adopt the iterative refinement techniques for more improved performance.

The iterative technique possesses inherent characteristic of iteratively converging towards the desired solution. In this study, we adopt this technique to perform effective multi-sound source localization. Though this approach, we automatically and iteratively identify multiple sound sources, enhancing the localization for the same objects and allowing differentiation between different objects. As a result, our method demonstrates improved performance compared to conventional methods. Moreover, due to its automatic object-finding nature, our iterative approach is not constrained by prior source knowledge.

%% file: sec/3_method.tex
\section{Proposed Approach}
\subsection{Overall Architecture}

Figure \ref{fig:2} shows the overall architecture of our framework. Similar to the previous works \cite{s_CVPR2018_Senocak, s_CVPR2021_lvs}, we adopt a two-stream network to extract visual and audio features. Input image set $I_v\in\mathbb{R}^{B\times W_v \times H_v \times 3}$ ($B$ indicates batch number, $W_v$ and $H_v$ denote width and height of $I_v$) and the corresponding mixed audio spectrogram $I_a\in\mathbb{R}^{B\times W_a\times H_a \times 1}$ ($W_a$ and $H_a$ denote width and height of $I_a$) pass through each modal encoder (\textit{i.e.,} visual and audio encoders) to generate visual feature $\textbf{F}_v$ and audio feature $\textbf{F}_a$, respectively. First, we identify all the sound-making regions associated with mixed audios to filter the regions of interest. We measure cosine similarity between $\textbf{F}_v$ and $l_a$, derived from the global average pooling (GAP) of $\textbf{F}_a$, to obtain a sound-associated map $\textbf{S}_v$. 

Next, we obtain a sound-associated visual feature $\hat{\textbf{F}}_{v}$ by the product of $\textbf{S}_v$ and $\textbf{F}_v$ to being aware of features for overall sound-making regions. The Iterative Object Identification (IOI) module receives $\hat{\textbf{F}}_{v}$ to iteratively discriminate sound-associated grid cells and collect each cell in an object bank $\textbf{O}$. Subsequently, sound-associated grid cells in $\textbf{O}$ are allocated into object groups (\textit{e.g.,} object 1, object 2, etc.), and the background regions are removed. Our method differentiates various objects without prior source knowledge (\textit{i.e.,} the number of sound sources).


\subsection{Sound-Associated Region Localization}
\label{sec:sarl}
To automatically localize sound-making objects in a visual scene without any annotation or prior knowledge, it is essential to measure the correspondence between audio and visual features. To achieve this, we try to distinguish sound-making regions from non-sound regions.

We calculate the cosine similarity between $\textbf{F}_v\in\mathbb{R}^{B\times w\times h\times c}$ ($w$, $h$, and $c$ are the width, height, and channel) and $l_a\in\mathbb{R}^{B\times c}$ to generate sound-associated map $\textbf{S}_v=\{S_{v_{ij}}\}_{i=1,...,h, j=1,...,w}\in\mathbb{R}^{B\times w \times h}$.  To make $\textbf{S}_v$ meaningful, our method allows  positive audio-visual pairs to be attracted to each other and negative pairs to be repelled at training time. By doing so, it is possible to identify spatial regions associated with given sound (positive pairs) and regions which are not associated with that sound (negative pairs). We define $\textbf{S}^{n \rightarrow m}_{v}$ as cosine similarity map between $n$-th image and $m$-th audio, and the $ij$-th element of $\textbf{S}^{n \rightarrow m}_{v}$, which can be represented as: 
\begin{equation}
\begin{gathered}
Sim(A,B) =\frac{\langle A,B\rangle} {{||A||}\,{||B||}},\\
S^{n \rightarrow m}_{v_{ij}} = \frac{Sim({\textbf{F}}^n_{v_{ij}},l^m_a)}{\sum_{i=1}^{h}\sum_{j=1}^{w} Sim({\textbf{F}}^n_{v_{ij}},l^n_a)},
\end{gathered}
\end{equation}
where  $\langle \cdot , \cdot \rangle$ denotes inner-product. Next, we obtain the mask of $n$-th clip $M^n$, represented as:
\begin{equation}
M^n=sigmoid(({S}^{n\rightarrow n}_v-\alpha)/\omega),
\end{equation}
where $\alpha$ and $\omega$ are the hyper-parameters. Based on $M^n$, we aim to localize all the sound-making regions associated with mixed audios. Thus, motivated by \cite{s_CVPR2021_lvs}, we employ an audio-visual contrastive loss $\mathcal{L}_{avc}$, which can be represented as:
\begin{equation}
\begin{gathered}
Pos^n = \frac{1}{|M^n|} \langle M^n, {S}^{n \rightarrow n}_v\rangle, \\
Neg^n = \frac{\langle \textbf{1} - M^n, S^{n \rightarrow n}_v\rangle}{|\textbf{1} - M^n|} + \frac{1}{hw}\sum_{n \neq m} \langle \textbf{1}, S^{n \rightarrow m}_v \rangle, \\
\mathcal{L}_{avc} = -\frac{1}{B} \sum_{n=1}^{B} \left[ \log \frac{\exp(Pos^n)}{\exp(Pos^n) + \exp(Neg^n)} \right].
\end{gathered}
\end{equation}
${Pos}^n$ represents the average vector of highly correlated regions between the audio-visual pairs in the $n$-th clip. In contrast, ${Neg}^n$ contains not only regions with low correlated between audio-visual pairs within the same clip but also includes correlation with other clips. Consequently, $\mathcal{L}_{avc}$ allows the $\textbf{S}_v$ to effectively filter out the foreground regions associated with the mixed audio, separating from the background regions that do not produce sound.

\subsection{Iterative Object Identification Module}
\label{sec:ioi}

Through Section \ref{sec:sarl}, the sound-associated map $\textbf{S}_v$ can contain knowledge of the sound-associated objects. We then multiply $\textbf{S}_v$ by the original visual feature $\textbf{F}_v$ to encode sound-associated visual feature $\hat{\textbf{F}}_v\in\mathbb{R}^{B\times w\times h\times c}$. By doing so, background cells are removed, leaving all the sound-associated cells that represent the sound-making object.

Next, the proposed Iterative Object Identification (IOI) module takes $\hat{\textbf{F}}_v$ to identify sound-making objects in mixed audio-visual data. Using $\hat{\textbf{F}}_v$, the IOI module conducts a total of $T$ iterations to find all sound-associated cells, and in each iteration, it finds the area to localize around the position of the sound-associated cell identified. Each iteration consists of four steps: (1) Highest sound-associated cell selection, (2) foreground/background selection, (3) finding sound-making regions, and (4) identified sound-making region exclusion. After the end of $T$ iteration process, we cluster all sound-associated grid cells as each object or negative. 

\setlength{\textfloatsep}{3pt}
\begin{algorithm}[t]
\caption{Iterative Object Identification Algorithm}
\label{alg:1}
\begin{algorithmic}
\small
\Require $\textbf{S}_v, \hat{\textbf{F}}_v$
\Ensure $\textbf{O}= [$ $]$, $t = 0$

\While{$\max(\textbf{S}_v) > \varepsilon$}
    \State $\textbf{E}^t_p \gets \hat{\textbf{F}}_{v}[ \text{argmax}(\textbf{S}_v)]$
    \State $\textbf{R}_p \gets \langle \hat{\textbf{F}}_v, \textbf{E}^t_p \rangle$ \Comment{Inner Product}
    \State $\textbf{R}_n \gets \langle \hat{\textbf{F}}_v, \textbf{E}_n \rangle$

    \State $sound\_making\_region \gets \textbf{R}_p > \textbf{R}_n$ 

    \State $\textbf{S}_v \gets \textbf{S}_v \cdot \neg sound\_making\_region$
    \State $\textbf{O}.append(\textbf{E}^t_p)$
    \State $t \gets t + 1$
\EndWhile
\end{algorithmic}
\small
\noindent{\textbf{Results: } {$T = t, \textbf{O} = [\textbf{E}^1_p, \textbf{E}^2_p, ..., \textbf{E}^T_p]$}}
\end{algorithm}
\setlength{\textfloatsep}{12pt}

Before starting iteration, we define a negative cell vector $\textbf{E}_n\in \mathbb{R}^{B \times c}$ to represent background (non-sound-making objects). $\textbf{E}_n$ is obtained by averaging the background cells that were excluded during the creation of $\hat{\textbf{F}}_v$. This vector is essential for identifying regions that do not correspond to the sound, to provide contrast to the foreground. 

\noindent{\textbf{(1) Highest sound-associated cell selection.}} In the first step in $t$-th iteration, we select the cell in ${\textbf{S}}_v$ with the highest value and select this cell as a starting point. This selection is based on the understanding that the highest value in ${\textbf{S}}_v$ indicates the strongest audio-visual correspondence in that cell. Therefore, we identify and mark the coordinates of this cell as $(i,j)$. The vector $\hat{\textbf{F}}_{v_{ij}}$ is then assigned as the sound-associated grid cell $\textbf{E}^t_p\in\mathbb{R}^{B\times c}$. Then, we store $\textbf{E}^t_p$ in the $i$-th element of the object bank $\textbf{O}=\{\textbf{E}^t_p\}^T_{t=1}$. 

\noindent{\textbf{(2) Foreground/background selection.}} In the second step in $t$-th iteration, we perform foreground and background selection using $\textbf{E}^t_p$ and $\textbf{E}_n$. The similarity measurement is carried out by calculating the inner product of $\hat{\textbf{F}}_v$ and $\textbf{E}^t_p$. This result is used as foreground region $\textbf{R}^t_p \in \mathbb{R}^{B \times w \times h}$. A higher value in $\textbf{R}^t_p$ indicates a higher likelihood of being the same object as $\textbf{E}^t_p$. Similarly, the background region is calculated by inner product of $\hat{\textbf{F}}_v$ and $\textbf{E}_n$, resulting in $\textbf{R}^t_n \in \mathbb{R}^{B\times w\times h}$.

\noindent{\textbf{(3) Finding sound-making regions.}} In the third step in $t$-th iteration, we compare $\textbf{R}^t_p$ and $\textbf{R}^t_n$ to localize sound-making regions. The choice of sound-making region sizes is based on the condition that the value of foreground region $\textbf{R}_p^t$ is higher than that of the background $\textbf{R}_n^t$. Using these regions, we generate the $t$-th localization map.

\begin{figure}[t]
    \begin{minipage}[b]{1.0\linewidth}
        \vspace{-0.1cm}
	\centering
	\centerline{\includegraphics[width=8.0cm]{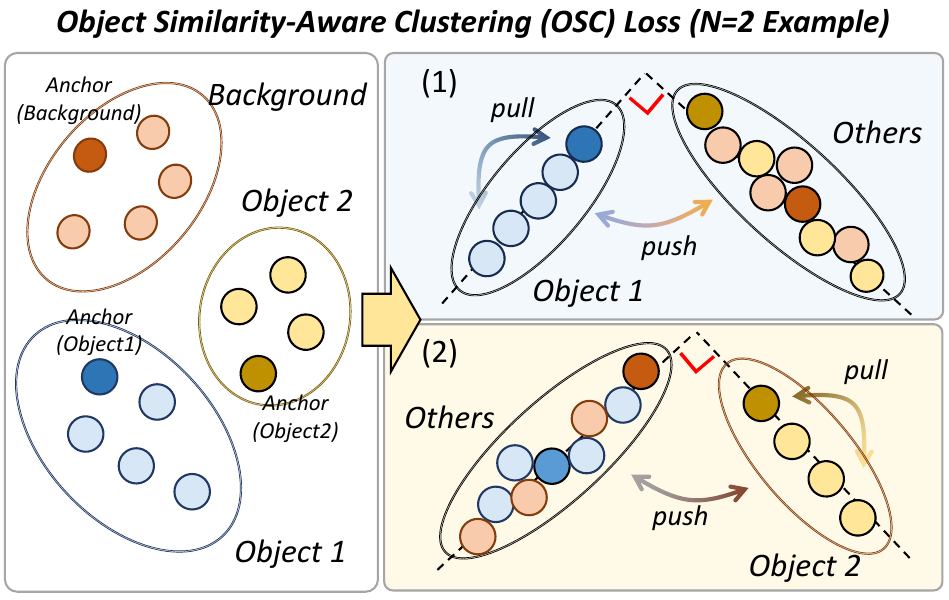}}
	\end{minipage}
        \vspace{-0.7cm}
	\caption{Explanation of the proposed Object Similarity-Aware Clustering (OSC) loss ($N=2$ example).}
    \vspace{-0.3cm}
    \label{fig:3}
\end{figure}

\noindent{\textbf{(4) Identified sound-making region exclusion.}} In the fourth step in $t$-th iteration, to prevent redundancy in finding sound-associated grid cell vector $\textbf{E}^t_p$, we need to modify $\textbf{S}_v$ for the next iteration. This involves setting the values of regions already localized to zero in $\textbf{S}_v$, thereby excluding them from further iteration. After completing these steps, we move on to the next iteration. We repeat this process until we find all sound-associated grid cells. The pseudocode detailing this procedure is presented in Algorithm \ref{alg:1}, offering a clear step-by-step guide.

\subsection{{Object Identification and Final Localization.}}
\label{sec:cluster}
After finishing the iteration process, we identify all sound-associated grid cells as each object or negative. Each cell is evaluated to determine whether it represents a sound-making object or background. This classification is crucial as it ensures the accuracy of our object identification.

First, to eliminate cells that were identified as sound-associated grid cells but are actually background, we calculate cosine similarity between negative cell vector $\textbf{E}_n$ and each sound-associated grid cell ${\textbf{E}^t_p}$ in object bank $\textbf{O}$. Cells with a similarity exceeding a threshold $\tau_1$ are considered to be a part of the background and are eliminated from $\textbf{O}$. 

Subsequently, to cluster the remaining sound-associated grid cells into each object, we calculate the cosine similarity between every pair of cells. This calculation helps us understand how similar or related these cells are to each other. If the similarity between any two cells is higher than the threshold $\tau_2$, we consider these cells to be part of the same object. To effectively cluster each object, we use the union-find algorithm \cite{union_find}. It merges other cells into each group if they are determined to belong to the same object. 

Each identified group is labeled as object $k$ (where $k = 1, 2, ..., K_b$). Here, $K_b$ is the number of objects in $b$-th clip. Finally, we combine the localization maps from object $k$ to create the final localization map for each object.

\subsection{Object Similarity-aware Clustering Loss}

To effectively identify and separate multi-sound sources without any annotation and prior source knowledge in complex environments, we propose an Object Similarity-aware Clustering (OSC) loss. The OSC loss has an important role in guiding the IOI module to combine regions belonging to the same object while differentiating between distinct objects and backgrounds. In the bank $\textbf{O}_b$, the anchor cell $A^b_{k}$ is identified as the most representative cell of the object $k$ in the $b$-th clip. $P^b_k$ and $N^b_k$ represent compositions of cells, where $P^b_k$ consists of cells attributed to the same object, and $N^b_k$ is from different objects or backgrounds ($P^b_k$ and $N^b_k$ are distinguished though Section \ref{sec:cluster}). The formulation of the OSC loss is structured as follows:  
\begin{equation}
\begin{gathered}
C^b_{pk} = Sim(A^b_{k} , P^b_k), \,\,\,\, C^b_{nk} = Sim(A^b_{k}, N^b_k), \\
\mathcal{L}^b_{k} = \left( 1 -C^b_{pk} \right) +C^b_{nk} ,\\
\mathcal{L}_{osc} = {\frac{1}{B}\sum_{b=1}^{B} \sum_{k=1}^{K_b} \frac{\mathcal{L}^b_{k}}{K_b}},
\end{gathered}
\end{equation}
\noindent where $C^b_{pk}$ and $C^b_{nk}$ represent the cosine similarity between the anchor and the positive, and the anchor and the negative, respectively. Through this loss, learning is directed in such a way that similarity between the anchor and positive cells is reinforced, while divergence between the anchor and negative cells is accentuated. As shown in Figure \ref{fig:3}, this effectively clusters cells that are considered to be part of the same object, thereby aiding in the segmentation of objects. This approach reduces the number of iterations to distinguish objects, as detailed in our supplementary experiments.

\subsection{Training Objective}
To train our proposed method, we construct the total training loss function as follows: 
\begin{equation}
\mathcal{L}_{Total}=\lambda_1\mathcal{L}_{avc}+\lambda_2\mathcal{L}_{osc},
\end{equation}
where $\lambda_1$ and $\lambda_2$ denote balancing parameters for each loss function. By using $\mathcal{L}_{Total}$, our method effectively performs multi-sound source localization, identifying sound sources from mixtures without the need for prior source knowledge.

\vspace{-0.15cm}

%% file: sec/4_experiments.tex
\begin{table*}[t]
    \renewcommand{\tabcolsep}{1.0mm}
    \centering
	\resizebox{0.95\linewidth}{!}{
		\begin{tabular}{ccccccccccc}
            \Xhline{3\arrayrulewidth}
            \rule{0pt}{10.0pt} \multirow{2}{*}{\bf Method}  & \multicolumn{5}{c}{\textbf{MUSIC-Duet} \cite{dataset_music}} & \multicolumn{5}{c}{\textbf{VGGSound-Duet} \cite{dataset_vggsound}} \\ 
            \cmidrule(lr){2-6} \cmidrule(l){7-11}
            & {\textbf{CAP(\%)}} & {\textbf{PIAP(\%)}} & {\textbf{CloU@0.3(\%)}} & {\textbf{AUC(\%)}} & {} & {\textbf{CAP(\%)}} & {\textbf{PIAP(\%)}} & {\textbf{CloU@0.3(\%)}} & {\textbf{AUC(\%)}} & {} \\
            \midrule
            Attention10k \cite{s_CVPR2018_Senocak} (CVPR'18) & {--} & {--} & 21.6 & 19.6 &  & {--} & {--} & 11.5 & 15.2 &  \\
            OTS \cite{s_2018ots} (ECCV'18) & 11.6 & 17.7 & 13.3 & 18.5 &  & 10.5 & 12.7 & 12.2 & 15.8 &  \\
            DMC \cite{s_hu2019dmc} (CVPR'19) & {--} & {--} & 17.5 & 21.1 &  & {--} & {--} & 13.8 & 17.1 &  \\
            CoarseToFIne \cite{m_ECCV2020_Qian_coarsetofine} (ECCV'20) & {--} & {--} & 17.6 & 20.6 &  & {--} & {--} & 14.7 & 18.5 &  \\
            DSOL \cite{m_hu2020discriminative} (NeurIPS'20) & {--} & {--} & 30.1 & 22.3 &  & {--} & {--} & 22.3 & 21.1 &  \\
            LVS \cite{s_CVPR2021_lvs} (CVPR'21) & {--} & {--} & 22.5 & 20.9 &  & {--} & {--} & 17.3 & 19.5 &  \\
            EZ-VSL \cite{s_mo2022easy} (ECCV'22) & {--} & {--} & 24.3 & 21.3 &  & {--} & {--} & 20.5 & 20.2 &  \\
            Mix-and-Localize \cite{m_hu2022mix} (CVPR'22) & 47.5 & 54.1 & 26.5 & 21.5 &  & 16.3 & 22.6 & 21.1 & 20.5 &  \\
            AVGN \cite{m_mo2023audio} (CVPR'23) & \underline{50.6} & \underline{57.2} & \underline{32.5} & \underline{24.6} &  & \underline{21.9} & \underline{28.1} & \underline{26.2} & \underline{23.8} &  \\\cdashline{1-10}
            \rule{0pt}{10.5pt}
            \bf Proposed Method & \bf 52.1 & \bf 72.5 & \bf 38.6 & \bf 30.1 &  & \bf 32.5 & \bf 44.4 & \bf 46.9 & \bf 29.2 &  \\
            \Xhline{3\arrayrulewidth}
            \end{tabular}
        }
    \vspace{-0.3cm}
    \caption{Experimental results on Music-Duet(left) and VGGSound-Duet(right) for multi-sound source localization. \textbf{Bold}/\underline{underlined} fonts indicate the best/second-best results.}
    \vspace{-0.05cm}
    \label{table:multi}
\end{table*}

\begin{table*}[t]
\small
    \renewcommand{\tabcolsep}{3.5mm}
    \centering\
	\resizebox{0.95\linewidth}{!}{
		\begin{tabular}{c ccc ccc}
            \Xhline{3\arrayrulewidth}
            \rule{0pt}{10.5pt} \multirow{2}{*}{\bf Method}  & \multicolumn{3}{c}{\textbf{MUSIC-Solo} \cite{dataset_music}}   & \multicolumn{3}{c}{\textbf{VGGSound-Single} \cite{dataset_vggsound}} \\ 
            \cmidrule(lr){2-4} \cmidrule(l){5-7}
            & {\textbf{AP(\%)}} & {\textbf{IoU@0.5(\%)}} & {\textbf{AUC(\%)}} & {\textbf{AP(\%)}} & {\textbf{IoU@0.5(\%)}} & {\textbf{AUC(\%)}} \\
            \midrule
            Attention10k \cite{s_CVPR2018_Senocak} (CVPR'18) & {--} & 37.2 & 38.7  & {--} & 19.2 & 30.6  \\
            OTS \cite{s_2018ots} (ECCV'18) & 69.3 & 26.1 & 35.8 &   29.8 & 32.8 & 35.7   \\
            DMC \cite{s_hu2019dmc} (CVPR'19) & {--} & 29.1 & 38.0   & {--} & 23.9 & 27.6   \\
            CoarseToFIne \cite{m_ECCV2020_Qian_coarsetofine} (ECCV'20) & 70.7 & 33.6 & 39.8   & 28.2 & 29.1 & 34.8   \\
            DSOL \cite{m_hu2020discriminative} (NeurIPS'20) & {--} & 51.4 & 43.7   & {--} & 35.7 & 37.2   \\
            LVS \cite{s_CVPR2021_lvs} (CVPR'21) & 70.6 & 41.9 & 40.3   & 29.6 & 34.4 & 38.2   \\
            EZ-VSL \cite{s_mo2022easy} (ECCV'22) & 71.5 & 45.8 & 41.2   & 31.3 & 38.9 & 39.5   \\
            Mix-and-Localize \cite{m_hu2022mix} (CVPR'22) & 68.6 & 30.5 & 40.8 &   32.5 & 36.3 & 38.9   \\
            AVGN \cite{m_mo2023audio} (CVPR'23) & \underline{77.2} & \underline{58.1} & \underline{48.5} &  \underline{35.3} & \underline{40.8} & \bf{42.3}  \\\cdashline{1-7}
            \rule{0pt}{10.5pt}
            \bf Proposed Method & \bf 77.4 & \bf 62.1 & \bf 59.4 &   \bf 46.2 & \bf 41.4 & \underline{41.2}    \\
            \Xhline{3\arrayrulewidth}
            \end{tabular}
        }
    \vspace{-0.3cm}
    \caption{Experimental results on Music-Solo(left) and VGGSound-Single(right) for single sound source localization. \textbf{Bold}/\underline{underlined} fonts indicate the best/second-best results.}
    \vspace{-0.5cm}
    \label{table:single}
\end{table*}

\section{Experiments}

\subsection{Datasets and Evaluation Metrics}
\noindent{\textbf{MUSIC.}} The MUSIC dataset \cite{dataset_music} comprises 448 unedited YouTube music videos featuring solos and duets across 11 categories of musical instruments. To ensure a fair comparison with previous works, we use the same training/testing subset as utilized in AVGN \cite{m_mo2023audio}. Specifically, the MUSIC-Solo contains 358 solo videos that are applied for training, and 90 solo videos are for evaluation for single sound source localization. Similarly, the MUSIC-Duet contains 124 duet videos that are applied for training, and 17 duet videos are for evaluation for multi-sound source localization. 

\begin{figure*}[t]
    \begin{minipage}[b]{1.0\linewidth}
				\centering
                \vspace{-0.6cm}
				\centerline{\includegraphics[width=17.4cm]{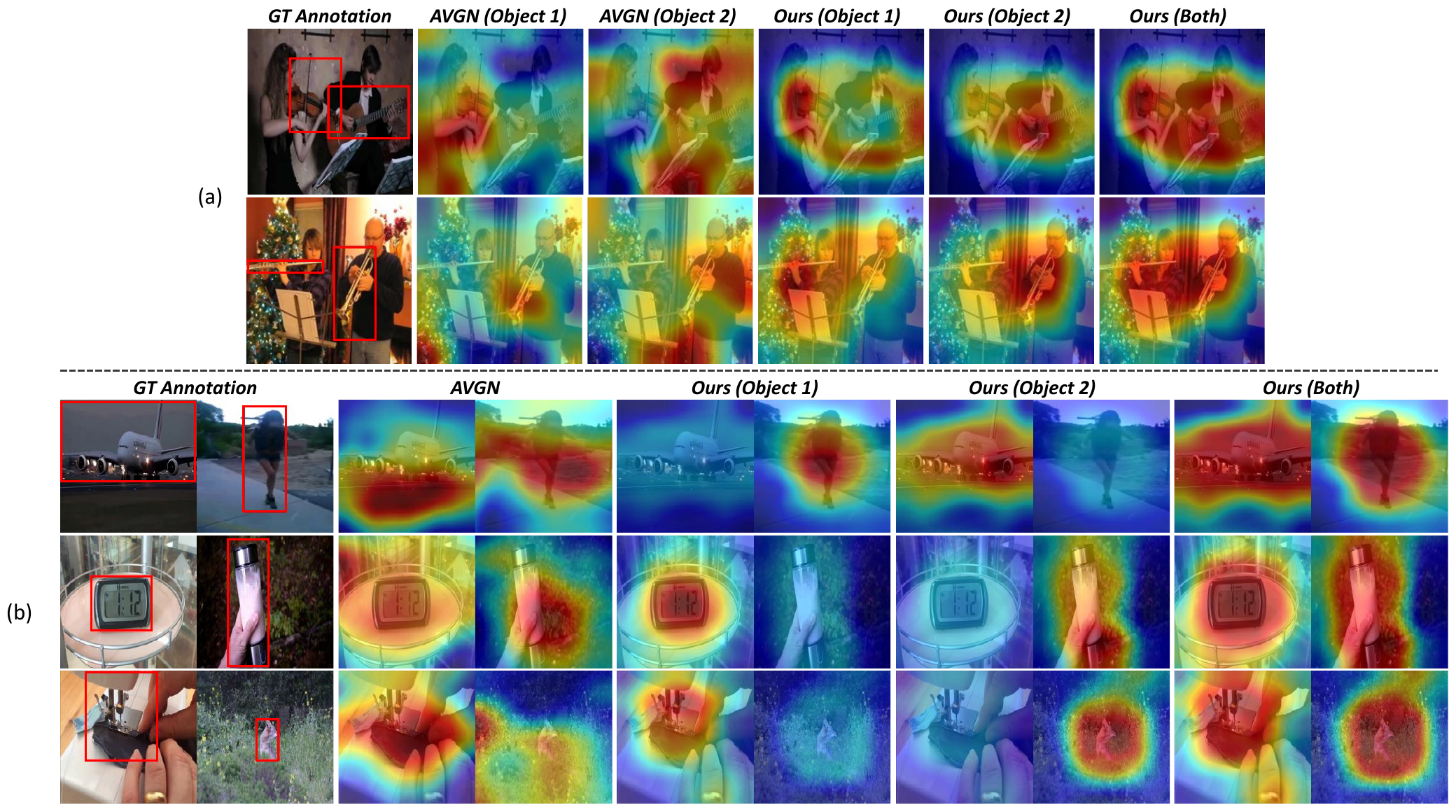}}
			\end{minipage}
			\vspace{-0.72cm}
			\caption{Visualization results for (a) MUSIC-Duet, (b) VGGSound-Duet test set. We compare our method with AVGN \cite{m_mo2023audio}.}
			\vspace{-0.3cm}
   \label{fig:4}
\end{figure*}

\begin{figure*}[t]
    \begin{minipage}[b]{1.0\linewidth}
				\centering
				\centerline{\includegraphics[width=15cm]{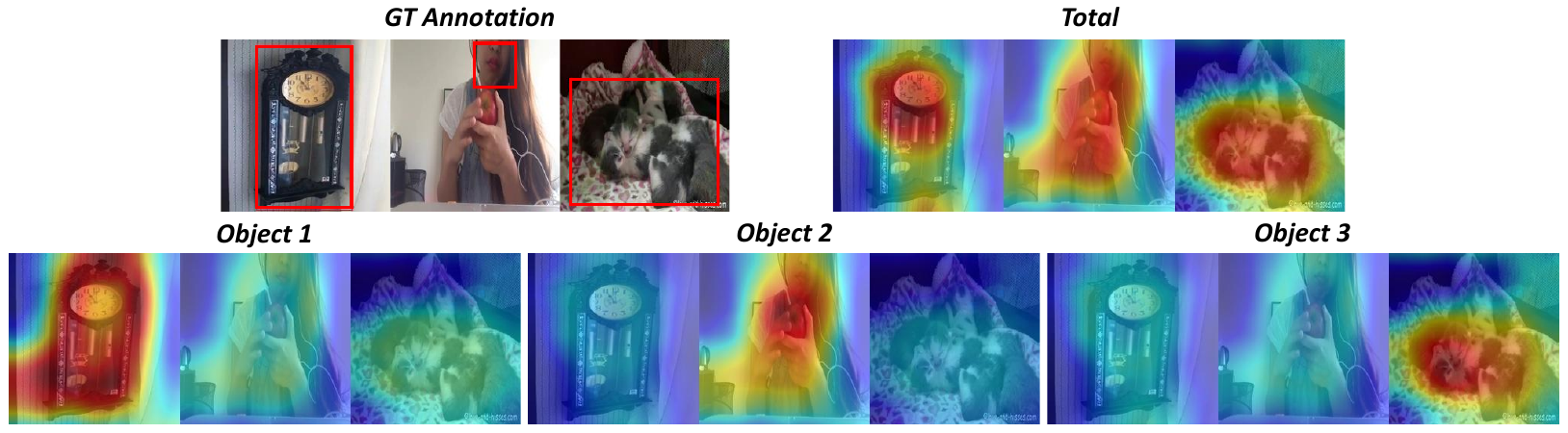}}
			\end{minipage}
			\vspace{-0.74cm}
			\caption{Visualization results of our method on VGGSound-Trio test set.}
			\vspace{-0.4cm}
   \label{fig:5}
		\end{figure*}

\noindent{\textbf{VGG-Sound.}} VGG-Sound dataset \cite{dataset_vggsound}, which is denoted as VGGSound-Single consists of more than 200k videos from 221 different sound categories. We utilize 144k image-audio pairs as a training set. For single sound source localization, we use VGG-Sound Source \cite{s_CVPR2021_lvs} to evaluate our method. In the training phase, for multi-sound source localization, we randomly concatenate two video frames to generate a single input image with dimensions of 448$\times$224, and we combine the corresponding audio waveforms to generate a mixed audio signal. Following \cite{m_mo2023audio}, we employ the VGGSound-Duet dataset for evaluation. 

\noindent{\textbf{Evaluation Metrics.}} Following the prior works \cite{m_hu2022mix, m_mo2023audio}, we adopt metrics for a fair and comprehensive comparison. For single sound source localization, we employ Average Precision (AP), Intersection over Union (IoU), and Area Under Curve (AUC). Regarding multi-sound source localization, we employ Class-aware Average Precision (CAP), Permutation-Invariant Average Precision (PIAP), Class-aware IoU (CIoU), and Area Under Curve (AUC). Our method is self-supervised and does not use class labels, we follow \cite{m_hu2022mix} for the modified version of CAP and CIoU.

\subsection{Implementation Details}
For the visual modality input, we resize images to dimensions of $W_v=224$, $H_v=224$, extracting them from the central frame of 3-second video clips. For the audio modality, we resample the raw 3-second audio signal to 16kHz and convert it into a log-scale spectrogram.

In accordance with the methods presented in \cite{s_CVPR2021_lvs}, we utilize a ResNet-18 \cite{resnet} for both the visual and audio feature backbones to establish our model. Due to the sound spectrogram having only one channel, we adapt the first convolutional layer of the ResNet-18 \cite{resnet} from 3 channels to 1. The visual encoder is pretrained using ImageNet \cite{deng2009imagenet}. Our method employs the Adam optimizer \cite{kingma2014adam} with a learning rate of $10^{-4}$ and a batch size of 128. We train our model for 50 epochs on all experiments. The training is conducted using 1 RTX 4090 GPU. We use the weights of $\tau_1 = 0.7$, $\tau_2 = 0.6$. For our proposed loss function $\mathcal{L}_{Total}$, we set $\lambda_1 = 1$ and $\lambda_2 = 1$. Following \cite{s_CVPR2021_lvs, s_WACV2023_htf}, we utilize the implementation code and adopt the same other hyper-parameters.

\subsection{Comparison to Prior Works}

\noindent{\textbf{Multi-Sound Source Localization.}} First, we conduct experiments on the multi-sound source localization scenarios. We compare our method with the state-of-the-art multi-sound source localization methods \cite{s_CVPR2018_Senocak, s_2018ots, s_hu2019dmc, m_ECCV2020_Qian_coarsetofine, m_hu2020discriminative, s_CVPR2021_lvs, s_mo2022easy, m_hu2022mix, m_mo2023audio}. Table \ref{table:multi} shows the results of our method on MUSIC-Duet \cite{dataset_music} and VGGSound-Duet \cite{dataset_vggsound}.
For the MUSIC-Duet test set, our method demonstrates superior performance, achieving 1.5\%, 15.3\%, 6.1\%, and 5.5\% higher for CAP, PIAP, CIoU@{0.3}, and AUC, respectively. Furthermore, in the VGGSound-Duet dataset, our method significantly outperforms existing methods. It indicates that our OSC loss enables our method to effectively localize sound-making objects through an iterative process.

\noindent{\textbf{Single Sound Source Localization.}} For single sound source localization, we compare our method with other existing works \cite{s_CVPR2018_Senocak, s_2018ots, s_hu2019dmc, m_ECCV2020_Qian_coarsetofine, m_hu2020discriminative, s_CVPR2021_lvs, s_mo2022easy, m_hu2022mix, m_mo2023audio}. Table \ref{table:single} presents the performances on MUSIC-Solo \cite{dataset_music} and VGGSound-Single datasets \cite{dataset_vggsound}. For the MUSIC-Solo test set, ours shows 0.2\% higher for AP. 4.0\% higher for IoU$@{0.5}$, and 10.9\% higher for AUC, respectively. For the VGGSound-Single test set, ours outperforms most of the others. For AUC, ours achieves the second-best performance.

The experimental results in Table \ref{table:multi} and Table \ref{table:single} clearly demonstrate the significant advancements our method brings to single and multi-sound source localization. This progress is largely attributed to the capabilities of the IOI module and OSC loss in effectively distinguishing objects.

\subsection{Visualization Results}

We compare ours with the state-of-the-art approach, AVGN \cite{m_mo2023audio}, in multi-sound source localization. This comparison utilizes the MUSIC-Duet and VGGSound-Duet test sets for visualizing the localization results. The results are shown in Figure \ref{fig:4}. The visualization highlights the efficacy of our method in accurately localizing sound-making objects within a mixture. Ground Truth (GT) annotations are used to denote regions where sound-making objects are present. 

Furthermore, Figure \ref{fig:5} offers additional visualization results from our method, showcasing its efficacy in differentiating objects in scenarios with three-source mixtures. We utilize VGGSound-Trio test set comprising a mixture of three sound sources from VGGSound-Single \cite{dataset_vggsound}, following \cite{m_mo2023audio}. Our IOI module facilitates the repeated detection and distinction of objects in the audio-visual scene, culminating in highly accurate and refined localization maps. 

\begin{table}[t]
    \centering
	\begin{center}
		\renewcommand{\tabcolsep}{1.0mm}
		\resizebox{0.999\linewidth}{!}
	       {
		  \begin{tabular}{c cc ccc}
                \Xhline{3\arrayrulewidth}
                \rule{0pt}{10pt} \textbf{Method} & $\mathcal{L}_{avc}$  & $\mathcal{L}_{osc}$ & \bf PIAP(\%) & \bf CloU@0.3(\%) & \bf AUC(\%) \\ \hline
                AVGN \cite{m_mo2023audio} & -       & - & 28.1 & 26.2 & 23.8 \\\cdashline{1-6}
                \rule{0pt}{9.5pt} \multirow{3}{*}{\makecell{\bf Proposed\\ \bf Method}}  & \cmark  & - & 44.1 & 25.3 & 28.3 \\
                & -       & \cmark & 44.3 & 43.0 & 27.0 \\
                & \cmark       & \cmark & \textbf{44.4}& \textbf{46.9} & \textbf{29.2} \\\Xhline{3\arrayrulewidth}
            \end{tabular}}
    \end{center}
    \vspace{-0.6cm}
    \caption{Effect of the proposed $\mathcal{L}_{avc}$ and $\mathcal{L}_{osc}$ loss on VGGSound-Duet test set.}
            \label{table:loss}
       \vspace{-0.2cm}     
\end{table}

\subsection{Ablation Study}
\noindent{\textbf{Effect of the Proposed Losses.}} We evaluate the performance impact of our two proposed losses $\mathcal{L}_{avc}$ and $\mathcal{L}_{OSC}$. The results are shown in Table \ref{table:loss}. When each loss is considered, our method shows the improved performance against AVGN \cite{m_mo2023audio} which is currently a state-of-the-art method. When all the proposed losses are taken into account, we show the highest performance. By integrating the proposed losses into our training, ours achieves increased capacity for learning robust and discriminative features.

\noindent{\textbf{Variation of $\tau_1$ and $\tau_2$.} We conduct an additional ablation study to investigate the effect of our method on the hyperparameters $\tau_1$ and $\tau_2$ as described in Section 3.3. The results in Table \ref{table:tau} indicate that we obtain the best results when $\tau_1=0.7$ and $\tau_2=0.6$. Importantly, even when $\tau_1$ and $\tau_2$ are varied, our method still outperforms the existing methods. These results suggest that the proposed model is robust to the variations of hyperparameters.

\begin{table}[t]

    \centering
	\begin{center}
		\renewcommand{\tabcolsep}{3.2mm}
            \vspace{-0.1cm}
		\resizebox{0.97\linewidth}{!}
	       {
		  \begin{tabular}{c cc cc}
                \Xhline{3\arrayrulewidth}
                $\tau_1$ & $\tau_2$ & \bf PIAP(\%) &\bf CloU@0.3(\%) & \bf AUC(\%) \\ \hline
                \multirow{3}{*}{\makecell{0.7}} & 0.6 & \bf44.4 & \bf46.9 & \bf29.2 \\
                
                & 0.7 & 44.1 & 43.9 & 27.5\\
                
                & 0.8 & 44.3 & 45.0 & 28.2 \\
                
                 \cdashline{1-5}
                
                0.6 & \multirow{3}{*}{\makecell{0.6}} & 44.3 & 46.3 & 28.9 \\
                0.7 &  & \bf44.4 & \bf46.9 & \bf29.2 \\
                0.8 &  & 44.2 & 46.8 & 29.1
                 \\\Xhline{3\arrayrulewidth}
                \end{tabular}}
    \end{center}
    \vspace{-0.65cm}
    \caption{Experimental results on VGGSound-Duet test set according to the hyper-parameters $\tau_1$ and $\tau_2$ for IOI module in Sec \ref{sec:cluster}.}
    \vspace{-0.3cm}
            \label{table:tau}
\end{table}

\subsection{Discussions}

\noindent{\textbf{Sound Source Counting Accuracy.}} To validate the accuracy of counting the correct sound source for our model, we employed each VGGSound dataset, which comprises images containing varying sound sources (ranging from one to three per image). Respectively, our model achieved an impressive accuracy of 90.79\%, 83.21\%, and 68.94\% for VGGSound-Single, Duet, and Trio test sets. These results not only highlight our performance in handling various numbers of sound sources but also demonstrate its potential applicability in diverse and dynamically changing audio-visual environments.

\noindent{\textbf{Adaptability of Our Method to Varying Sound Sources.}} We present experiments designed to evaluate the adaptability of our method in scenarios with varied sound sources. For this purpose, we utilized two distinct test sets: VGGSound-Trio \cite{m_mo2023audio} and VGGSound-Mixed. VGGSound-Mixed is a resampled dataset encompassing sounds from VGGSound-Single, VGGSound-Duet, and VGGSound-Trio. VGGSound-Mixed allows us to demonstrate the capability of our method to localize sound sources effectively, regardless of their quantity. The results are shown in Table \ref{table:mixed}. For the VGGSound-Trio test set, our method outperforms AVGN \cite{m_mo2023audio} across all evaluation metrics. For the VGGSound-Mixed test set, our approach achieves scores of 27.9\%, 36.0\%, 35.2\%, and 22.7\% for CAP, PIAP, CIoU@{0.3}, and AUC, respectively. Note that AVGN requires fixed prior source information, which limits its applicability. These results highlight our effectiveness in object localization, especially without prior source information, attributable to the efficacy of IOI module and loss function.

\noindent{\textbf{Computational Costs.}} Table \ref{table:cost} presents comparisons of training time, inference time, and the number of parameters. We compare our method with AVGN \cite{m_mo2023audio} which shows the highest performance among the existing methods. Due to the iterative method, our training time is marginally increased (7.14\%). However, inference time decreased by 28.3\%, and the number of parameters significantly decreased compared to the existing method with transformers.

\begin{table}[t!]
    \centering
	\begin{center}
		\renewcommand{\tabcolsep}{0.2mm}
		\resizebox{0.999\linewidth}{!}
	    {
		  \begin{tabular}{c c cccc}
                \Xhline{3\arrayrulewidth}
                \rule{0pt}{10pt} \textbf{Method} & \bf Test Set & \bf CAP(\%)& \bf PIAP(\%) & \bf CloU@0.3(\%) & \bf AUC(\%) \\ \hline 
                
                AVGN \cite{m_mo2023audio}    & \multirow{2}{*}{\makecell{\bf VGGSound-Trio}}   & 18.5 & 23.7 & 22.7 & 21.8 \\
                
                \bf Ours && \textbf{29.0} & \textbf{42.1}& \textbf{34.0} & \textbf{29.3}
                
                \\\cdashline{1-6}
                
                AVGN \cite{m_mo2023audio}    & \multirow{2}{*}{\makecell{\bf VGGSound-Mixed}}   & N/A & 22.9 & N/A & N/A \\
                
                \bf Ours && \textbf{27.9} & \textbf{36.0}& \textbf{35.2} & \textbf{22.7} 
                \\\Xhline{3\arrayrulewidth}
                \end{tabular}
            }
        \vspace{-0.3cm}
        \caption{Experimental results on VGGSound-Trio and VGGSound -Mixed test set. `$N/A$' denotes Not Available.}
        \label{table:mixed}
    \end{center}
    \vspace{-0.4cm}
\end{table}

		\begin{table}[t]
			\renewcommand{\tabcolsep}{0.7mm}
			\centering
			\resizebox{0.96\linewidth}{!}{
				\begin{tabular}{cccc}
					\Xhline{3\arrayrulewidth}
					\rule{0pt}{9.5pt} \bf \multirow{2}{*}{\bf Method} & \bf Training (s) & \bf Inference (s) & \multirow{2}{*}{\bf \#params}\\
					& \bf (\textit{per iter}) & \bf (\textit{per image})
					\\ \hline
					\rule{0pt}{10.pt}
					AVGN \cite{m_mo2023audio} (CVPR'23)      & 0.42 & 0.034 & 61.2M  \\
					\textbf{Proposed Method} & 0.45 & 0.027 & 38.5M \\
					\Xhline{3\arrayrulewidth}
				\end{tabular}
			}
            \vspace{-0.3cm}
            \caption{The comparisons of training time, inference time, and the number of parameters.}

            \vspace{-0.4cm}
			\label{table:cost}
		\end{table}

%% file: main.bbl
\begin{thebibliography}{48}
\providecommand{\natexlab}[1]{#1}
\providecommand{\url}[1]{\texttt{#1}}
\expandafter\ifx\csname urlstyle\endcsname\relax
  \providecommand{\doi}[1]{doi: #1}\else
  \providecommand{\doi}{doi: \begingroup \urlstyle{rm}\Url}\fi

\bibitem[Adhikari and Huttunen(2021)]{i_object_detection2}
Bishwo Adhikari and Heikki Huttunen.
\newblock Iterative bounding box annotation for object detection.
\newblock In \emph{ICPR}, 2021.

\bibitem[Aho and Hopcroft(1974)]{union_find}
Alfred~V Aho and John~E Hopcroft.
\newblock \emph{The design and analysis of computer algorithms}.
\newblock Pearson Education India, 1974.

\bibitem[Arandjelovic and Zisserman(2018)]{s_2018ots}
Relja Arandjelovic and Andrew Zisserman.
\newblock Objects that sound.
\newblock In \emph{ECCV}, 2018.

\bibitem[Bae et~al.(2022)Bae, Budvytis, and Cipolla]{i_single_view_depth}
Gwangbin Bae, Ignas Budvytis, and Roberto Cipolla.
\newblock Irondepth: Iterative refinement of single-view depth using surface normal and its uncertainty.
\newblock \emph{arXiv preprint arXiv:2210.03676}, 2022.

\bibitem[Biemond et~al.(1990)Biemond, Lagendijk, and Mersereau]{i_image_deblurring2}
Jan Biemond, Reginald~L Lagendijk, and Russell~M Mersereau.
\newblock Iterative methods for image deblurring.
\newblock \emph{Proceedings of the IEEE}, 1990.

\bibitem[Busso et~al.(2005)Busso, Hernanz, Chu, Kwon, Lee, Georgiou, Cohen, and Narayanan]{busso2005smart_speaker}
Carlos Busso, Sergi Hernanz, Chi-Wei Chu, Soon-il Kwon, Sung Lee, Panayiotis~G Georgiou, Isaac Cohen, and Shrikanth Narayanan.
\newblock Smart room: Participant and speaker localization and identification.
\newblock In \emph{ICASSP}, 2005.

\bibitem[Chen et~al.(2020)Chen, Xie, Vedaldi, and Zisserman]{dataset_vggsound}
Honglie Chen, Weidi Xie, Andrea Vedaldi, and Andrew Zisserman.
\newblock Vggsound: A large-scale audio-visual dataset.
\newblock In \emph{ICASSP}, 2020.

\bibitem[Chen et~al.(2021)Chen, Xie, Afouras, Nagrani, Vedaldi, and Zisserman]{s_CVPR2021_lvs}
Honglie Chen, Weidi Xie, Triantafyllos Afouras, Arsha Nagrani, Andrea Vedaldi, and Andrew Zisserman.
\newblock Localizing visual sounds the hard way.
\newblock In \emph{CVPR}, 2021.

\bibitem[Chen and He(2021)]{s_chen2021exploring}
Xinlei Chen and Kaiming He.
\newblock Exploring simple siamese representation learning.
\newblock In \emph{CVPR}, 2021.

\bibitem[Deng et~al.(2009)Deng, Dong, Socher, Li, Li, and Fei-Fei]{deng2009imagenet}
Jia Deng, Wei Dong, Richard Socher, Li-Jia Li, Kai Li, and Li Fei-Fei.
\newblock Imagenet: A large-scale hierarchical image database.
\newblock In \emph{CVPR}, 2009.

\bibitem[Fedorishin et~al.(2023)Fedorishin, Mohan, Jawade, Setlur, and Govindaraju]{s_WACV2023_htf}
Dennis Fedorishin, Deen~Dayal Mohan, Bhavin Jawade, Srirangaraj Setlur, and Venu Govindaraju.
\newblock Hear the flow: Optical flow-based self-supervised visual sound source localization.
\newblock In \emph{WACV}, 2023.

\bibitem[Hawley et~al.(1999)Hawley, Litovsky, and Colburn]{hawley1999speech}
Monica~L Hawley, Ruth~Y Litovsky, and H~Steven Colburn.
\newblock Speech intelligibility and localization in a multi-source environment.
\newblock \emph{JASA}, 1999.

\bibitem[He et~al.(2016)He, Zhang, Ren, and Sun]{resnet}
Kaiming He, Xiangyu Zhang, Shaoqing Ren, and Jian Sun.
\newblock Deep residual learning for image recognition.
\newblock In \emph{CVPR}, 2016.

\bibitem[Hoshiba et~al.(2017)Hoshiba, Washizaki, Wakabayashi, Ishiki, Kumon, Bando, Gabriel, Nakadai, and Okuno]{hoshiba2017_uav}
Kotaro Hoshiba, Kai Washizaki, Mizuho Wakabayashi, Takahiro Ishiki, Makoto Kumon, Yoshiaki Bando, Daniel Gabriel, Kazuhiro Nakadai, and Hiroshi~G Okuno.
\newblock Design of uav-embedded microphone array system for sound source localization in outdoor environments.
\newblock \emph{Sensors}, 2017.

\bibitem[Hu et~al.(2019)Hu, Nie, and Li]{s_hu2019dmc}
Di Hu, Feiping Nie, and Xuelong Li.
\newblock Deep multimodal clustering for unsupervised audiovisual learning.
\newblock In \emph{CVPR}, 2019.

\bibitem[Hu et~al.(2020)Hu, Qian, Jiang, Tan, Wen, Ding, Lin, and Dou]{m_hu2020discriminative}
Di Hu, Rui Qian, Minyue Jiang, Xiao Tan, Shilei Wen, Errui Ding, Weiyao Lin, and Dejing Dou.
\newblock Discriminative sounding objects localization via self-supervised audiovisual matching.
\newblock In \emph{NeurIPS}, 2020.

\bibitem[Hu et~al.(2022)Hu, Chen, and Owens]{m_hu2022mix}
Xixi Hu, Ziyang Chen, and Andrew Owens.
\newblock Mix and localize: Localizing sound sources in mixtures.
\newblock In \emph{CVPR}, 2022.

\bibitem[Kingma and Ba(2014)]{kingma2014adam}
Diederik~P Kingma and Jimmy Ba.
\newblock Adam: A method for stochastic optimization.
\newblock \emph{arXiv preprint arXiv:1412.6980}, 2014.

\bibitem[Li et~al.(2019)Li, Chen, Zhang, Liu, Xia, Cao, Young, and Xu]{i_image_deblurring1}
Taihao Li, Huai Chen, Min Zhang, Shupeng Liu, Shunren Xia, Xinhua Cao, Geoffrey~S Young, and Xiaoyin Xu.
\newblock A new design in iterative image deblurring for improved robustness and performance.
\newblock \emph{Pattern Recognition}, 2019.

\bibitem[Li et~al.(2016)Li, Girin, Badeig, and Horaud]{li2016robotics}
Xiaofei Li, Laurent Girin, Fabien Badeig, and Radu Horaud.
\newblock Reverberant sound localization with a robot head based on direct-path relative transfer function.
\newblock In \emph{IROS}, 2016.

\bibitem[Lin et~al.(2023)Lin, Tseng, Lee, Lin, and Yang]{s_iterative2023}
Yan-Bo Lin, Hung-Yu Tseng, Hsin-Ying Lee, Yen-Yu Lin, and Ming-Hsuan Yang.
\newblock Unsupervised sound localization via iterative contrastive learning.
\newblock \emph{CVIU}, 2023.

\bibitem[Liu et~al.(2022)Liu, Ju, Xie, and Zhang]{s_liu2022exploiting}
Jinxiang Liu, Chen Ju, Weidi Xie, and Ya Zhang.
\newblock Exploiting transformation invariance and equivariance for self-supervised sound localisation.
\newblock In \emph{ACM MM}, 2022.

\bibitem[Ma et~al.(2020)Ma, Jiang, Rao, Lu, and Zhou]{i_face_super}
Cheng Ma, Zhenyu Jiang, Yongming Rao, Jiwen Lu, and Jie Zhou.
\newblock Deep face super-resolution with iterative collaboration between attentive recovery and landmark estimation.
\newblock In \emph{CVPR}, 2020.

\bibitem[Mo and Morgado(2022{\natexlab{a}})]{s_mo2022easy}
Shentong Mo and Pedro Morgado.
\newblock Localizing visual sounds the easy way.
\newblock In \emph{ECCV}, 2022{\natexlab{a}}.

\bibitem[Mo and Morgado(2022{\natexlab{b}})]{s_momentum}
Shentong Mo and Pedro Morgado.
\newblock A closer look at weakly-supervised audio-visual source localization.
\newblock In \emph{NeurIPS}, 2022{\natexlab{b}}.

\bibitem[Mo and Tian(2023)]{m_mo2023audio}
Shentong Mo and Yapeng Tian.
\newblock Audio-visual grouping network for sound localization from mixtures.
\newblock In \emph{CVPR}, 2023.

\bibitem[Qian et~al.(2020)Qian, Hu, Dinkel, Wu, Xu, and Lin]{m_ECCV2020_Qian_coarsetofine}
Rui Qian, Di Hu, Heinrich Dinkel, Mengyue Wu, Ning Xu, and Weiyao Lin.
\newblock Multiple sound sources localization from coarse to fine.
\newblock In \emph{ECCV}, 2020.

\bibitem[Rascon and Meza(2017)]{rascon2017robotics}
Caleb Rascon and Ivan Meza.
\newblock Localization of sound sources in robotics: A review.
\newblock \emph{Robotics and Autonomous Systems}, 2017.

\bibitem[Saharia et~al.(2022)Saharia, Ho, Chan, Salimans, Fleet, and Norouzi]{i_super_resolution}
Chitwan Saharia, Jonathan Ho, William Chan, Tim Salimans, David~J Fleet, and Mohammad Norouzi.
\newblock Image super-resolution via iterative refinement.
\newblock \emph{TPAMI}, 2022.

\bibitem[Salvati et~al.(2019)Salvati, Drioli, Ferrin, and Foresti]{salvati2019_uav}
Daniele Salvati, Carlo Drioli, Giovanni Ferrin, and Gian~Luca Foresti.
\newblock Acoustic source localization from multirotor uavs.
\newblock \emph{TIE}, 2019.

\bibitem[Sayoud et~al.(2011)Sayoud, Ouamour, and Khennouf]{sayoud2011speaker}
Halim Sayoud, Siham Ouamour, and Salah Khennouf.
\newblock Speaker localization using stereo-based sound source localization.
\newblock In \emph{WOSSPA}, 2011.

\bibitem[Senocak et~al.(2018)Senocak, Oh, Kim, Yang, and Kweon]{s_CVPR2018_Senocak}
Arda Senocak, Tae-Hyun Oh, Junsik Kim, Ming-Hsuan Yang, and In~So Kweon.
\newblock Learning to localize sound source in visual scenes.
\newblock In \emph{CVPR}, 2018.

\bibitem[Senocak et~al.(2022)Senocak, Ryu, Kim, and Kweon]{s_hard_positive_mining}
Arda Senocak, Hyeonggon Ryu, Junsik Kim, and In~So Kweon.
\newblock Learning sound localization better from semantically similar samples.
\newblock In \emph{ICASSP}, 2022.

\bibitem[Senocak et~al.(2023)Senocak, Ryu, Kim, Oh, Pfister, and Chung]{s_senocak2023sound}
Arda Senocak, Hyeonggon Ryu, Junsik Kim, Tae-Hyun Oh, Hanspeter Pfister, and Joon~Son Chung.
\newblock Sound source localization is all about cross-modal alignment.
\newblock In \emph{ICCV}, 2023.

\bibitem[Shi and Ma(2022)]{s_WACV2022_Shi}
Jiayin Shi and Chao Ma.
\newblock Unsupervised sounding object localization with bottom-up and top-down attention.
\newblock In \emph{WACV}, 2022.

\bibitem[Singh et~al.(2023)Singh, Zinemanas, Serra, Bello, and Fuentes]{s_flowgrad}
Rajsuryan Singh, Pablo Zinemanas, Xavier Serra, Juan~Pablo Bello, and Magdalena Fuentes.
\newblock Flowgrad: Using motion for visual sound source localization.
\newblock In \emph{ICASSP}, 2023.

\bibitem[Sofiiuk et~al.(2022)Sofiiuk, Petrov, and Konushin]{i_segmentation2}
Konstantin Sofiiuk, Ilya~A Petrov, and Anton Konushin.
\newblock Reviving iterative training with mask guidance for interactive segmentation.
\newblock In \emph{ICIP}, 2022.

\bibitem[Song et~al.(2022)Song, Wang, Fan, Tan, and Zhang]{s_song2022sspl}
Zengjie Song, Yuxi Wang, Junsong Fan, Tieniu Tan, and Zhaoxiang Zhang.
\newblock Self-supervised predictive learning: A negative-free method for sound source localization in visual scenes.
\newblock In \emph{CVPR}, 2022.

\bibitem[Sun et~al.(2023)Sun, Zhang, Wang, Liu, Zhong, Feng, Guo, Zhang, and Barnes]{s_sun2023learning}
Weixuan Sun, Jiayi Zhang, Jianyuan Wang, Zheyuan Liu, Yiran Zhong, Tianpeng Feng, Yandong Guo, Yanhao Zhang, and Nick Barnes.
\newblock Learning audio-visual source localization via false negative aware contrastive learning.
\newblock In \emph{CVPR}, 2023.

\bibitem[Tian et~al.(2018)Tian, Shi, Li, Duan, and Xu]{s_crossmodal}
Yapeng Tian, Jing Shi, Bochen Li, Zhiyao Duan, and Chenliang Xu.
\newblock Audio-visual event localization in unconstrained videos.
\newblock In \emph{ECCV}, 2018.

\bibitem[Um et~al.(2023)Um, Kim, and Kim]{s_um2023sira}
Sung~Jin Um, Dongjin Kim, and Jung~Uk Kim.
\newblock Audio-visual spatial integration and recursive attention for robust sound source localization.
\newblock In \emph{ACM MM}, 2023.

\bibitem[Wang et~al.(2019)Wang, Shen, Cheng, and Shao]{i_object_detection1}
Wenguan Wang, Jianbing Shen, Ming-Ming Cheng, and Ling Shao.
\newblock An iterative and cooperative top-down and bottom-up inference network for salient object detection.
\newblock In \emph{CVPR}, 2019.

\bibitem[Wei et~al.(2023)Wei, Zhang, Ren, Li, Fu, and Xue]{i_sfm}
Xingkui Wei, Yinda Zhang, Xinlin Ren, Zhuwen Li, Yanwei Fu, and Xiangyang Xue.
\newblock Deepsfm: Robust deep iterative refinement for structure from motion.
\newblock \emph{TPAMI}, 2023.

\bibitem[Xuan et~al.(2022)Xuan, Wu, Yang, Yan, and Alameda-Pineda]{s_CVPR2022_ppsl}
Hanyu Xuan, Zhiliang Wu, Jian Yang, Yan Yan, and Xavier Alameda-Pineda.
\newblock A proposal-based paradigm for self-supervised sound source localization in videos.
\newblock In \emph{CVPR}, 2022.

\bibitem[Zhang et~al.(2019)Zhang, Lin, Liu, Yao, and Shen]{i_segmentation3}
Chi Zhang, Guosheng Lin, Fayao Liu, Rui Yao, and Chunhua Shen.
\newblock Canet: Class-agnostic segmentation networks with iterative refinement and attentive few-shot learning.
\newblock In \emph{CVPR}, 2019.

\bibitem[Zhao et~al.(2018)Zhao, Gan, Rouditchenko, Vondrick, McDermott, and Torralba]{dataset_music}
Hang Zhao, Chuang Gan, Andrew Rouditchenko, Carl Vondrick, Josh McDermott, and Antonio Torralba.
\newblock The sound of pixels.
\newblock In \emph{ECCV}, 2018.

\bibitem[Zhen et~al.(2020)Zhen, Wang, Zhou, Li, Shen, Shang, Fang, and Quan]{i_segmentation1}
Mingmin Zhen, Jinglu Wang, Lei Zhou, Shiwei Li, Tianwei Shen, Jiaxiang Shang, Tian Fang, and Long Quan.
\newblock Joint semantic segmentation and boundary detection using iterative pyramid contexts.
\newblock In \emph{CVPR}, 2020.

\bibitem[Zhou et~al.(2023)Zhou, Zhou, Hu, Zhou, and Ouyang]{s_WACV2023_Zhou}
Xinchi Zhou, Dongzhan Zhou, Di Hu, Hang Zhou, and Wanli Ouyang.
\newblock Exploiting visual context semantics for sound source localization.
\newblock In \emph{WACV}, 2023.

\end{thebibliography}
